\documentclass[11pt,onecolumn]{IEEEtran}

\usepackage{subfigure}
\usepackage{citesort}
\usepackage[fleqn]{amsmath}
\usepackage{ascmac}
\usepackage{afterpage}
\usepackage{cite}
\usepackage[dvips]{graphicx}
\usepackage{psfrag}
\usepackage{amsmath}
\usepackage{amsbsy}
\usepackage{amssymb}
\usepackage{latexsym}
\usepackage{color}

\graphicspath{{./images/}}

\newtheorem{proposition}{Proposition}
\newtheorem{example}{Example}

\newtheorem{theorem}{Theorem}
\newtheorem{lemma}{Lemma}
\newtheorem{corollary}{Corollary}

\newcommand{\migip}{\hspace*{\fill} $\Box $}

\newcommand{\euclidspace}{{\mathcal{H}}}
\newcommand{\signal}[1]{{\boldsymbol{#1}}}
\newcommand{\Natural}{{\mathbb N}}
\newcommand{\norm}[1]{\left\|#1\right\|}
\newcommand{\abs}[1]{\left|#1\right|}

\newcommand{\real}{{\mathbb R}}

\newcommand{\innerprod}[2]{\left\langle{#1},{#2}\right\rangle}

\newcommand{\refeq}[1]{(\ref{#1})}

\newcommand{\sinq}[1]{`#1'}

\newcommand{\argmin}{\operatornamewithlimits{argmin}}

\definecolor{darkgreen}{rgb}{0,.6,0}
\definecolor{medorange}{rgb}{0.7,0.3,0}
\definecolor{cyancyan}{rgb}{0.68, 0.92, 0.92}

\begin{document}
%
\title{Adaptive Learning in Cartesian Product of Reproducing Kernel Hilbert Spaces\vspace*{1em}}
%
%

 \author{Masahiro~Yukawa\\
Keio University, Dept.~Electronics and Electrical
        Engineering,\\
Hiyoshi 3-14-1, Kohoku-ku, Yokohama, Kanagawa, 223-8522 JAPAN\\
yukawa@elec.keio.ac.jp\\[5em]
\thanks{This work was partially supported by KDDI Foundation and
JSPS Grants-in-Aid (24760292).}
}

\markboth{}
{}
%



\maketitle

\begin{abstract}
We propose a novel adaptive learning algorithm based on
iterative orthogonal projections
in the Cartesian product of
multiple reproducing kernel Hilbert spaces (RKHSs).
The task is estimating/tracking nonlinear functions
which are supposed to contain multiple components
such as (i) linear and nonlinear components,
(ii) high- and low- frequency components etc.
In this case, the use of multiple RKHSs permits
a compact representation of multicomponent functions.
The proposed algorithm is where
two different methods of the author meet:
multikernel adaptive filtering and the algorithm of
hyperplane projection along affine subspace (HYPASS).
In a certain particular case,
the \sinq{sum} space of the RKHSs is isomorphic to the product space
and hence
the proposed algorithm can also be regarded as
an iterative projection method in the sum space.
The efficacy of the proposed algorithm is shown by numerical examples.
\end{abstract}

\begin{keywords}
reproducing kernel Hilbert space,
multikernel adaptive filtering,
Cartesian product,
orthogonal projection
\end{keywords}

%

\section{Introduction}\label{sec:intro}

Using reproducing kernels for nonlinear adaptive filtering tasks
has widely been investigated
\cite{kivinen04,engel04,huang_icassp05,liu08,slavakis08,slavakis09,richard09,liu_book10,thslya_IEEESPMAG11,vaerenbergh12,slavakis_ersp}.
See, e.g., \cite{aronszajn50,saitoh97,vapnik98,mueller01,schoelkopf01,berlinet04,shawe-taylor04,theodoridis08,spmag13,motai14} for the theory and applications of reproducing kernels.
The author has proposed and studied {\it multikernel adaptive filtering},
using \sinq{multiple} kernels
\cite{yukawa_sip10_multi,yukawa_eusipco11,yukawa_tsp12}.
Different approaches using multiple kernels have also been proposed subsequently.
Pokharel {\it et al.} have proposed a mixture-kernel approach
\cite{pokharel13},
and Gao {\it et al.} have proposed convex-combinations of kernel adaptive
filters \cite{gao_mlsp14}.
Tobar {\it et al.} have proposed a multikernel least mean square
algorithm for vector-valued functions \cite{tobar14}.
Multikernel adaptive filtering is effective particularly
in the following situations.
\begin{enumerate}
 \item[(a)] The unknown system to be estimated contains multiple
	    components with different characteristics such as
(i) linear and nonlinear components and 
(ii) high- and low- frequency components.
See \cite{haerdle00,espinoza05,xu_ac09,ylxu13}.
 \item[(b)] An adequate kernel is unavailable because 
(i) the amount of prior information about the unknown system is limited,
and/or
(ii) the unknown system is time-varying and so is 
the adequate kernel for the system.
\end{enumerate}
The situation (b) has mainly been supposed
in \cite{yukawa_sip10_multi,yukawa_eusipco11,yukawa_tsp12}.
Use of many, say fifty, kernels has been investigated
and kernel-dictionary joint-refinement techniques have been proposed
based on double regularization with a pair of block $\ell_1$ norms
\cite{yukawa_eusipco13,yukawa_apsipa13}.
Our primal focus in the current study is on the situation (a)
in which the use of multiple kernels is expected to
allow a compact representation of the unknown system.


Separately from the study of multikernel adaptive filtering,
the author has proposed
an efficient single-kernel adaptive filtering algorithm 
named hyperplane projection along affine subspace (HYPASS)
\cite{hypass,takizawa_tsp14}.
The HYPASS algorithm is a natural extension of 
the naive online $R_{\rm reg}$ minimization algorithm (NORMA)
proposed by Kivinen {\it et al.}~\cite{kivinen04}.
NORMA seeks to minimize a risk functional in terms of a nonlinear function
by using the stochastic gradient descent method
in a reproducing kernel Hilbert space (RKHS).
This approach builds a dictionary (the set of basic nonlinear functions
to generate an estimate of the unknown system)
by using all the observed data.
This implies that the dictionary size grows with
the number of data observed.
As a remedy for this issue,
a simple truncation rule has been introduced \cite{kivinen04}.
It would be more realistic to build a dictionary in a selective manner
based on some criterion to evaluate the novelty of a new datum;
simple criteria include Platt's criterion \cite{platt91},
the approximate linear dependency \cite{engel04}, and
the coherence criterion \cite{richard09}.
Introducing one of those criteria to NORMA raises another issue:
if a new datum is regarded to be not sufficiently novel 
and does not enter into the dictionary, then
this observed datum is simply discarded and makes no contributions to estimation
even though it can be informative enough to adjust the coefficients.
Moreover, the coefficient of each dictionary element is updated only when
that element enters into the dictionary.
The HYPASS algorithm systematically eliminates this limitation by enforcing
the update direction to lie in {\it the dictionary subspace} which is
spanned by the dictionary elements.
It has been extended to a parallel-projection-based algorithm
\cite{takizawa_icassp13,takizawa_tsp14}.
HYPASS includes the method of Dodd {\it et al.} \cite{dodd03} and
the quantized kernel LMS (QKLMS) \cite{chen_TNNLS12} as its particular case.
There are a similarity, and also a considerable dissimilarity,
between HYPASS and the kernel normalized least mean square (KNLMS)
algorithm \cite{richard09} proposed by Richard {\it et al.}
Both algorithms share the philosophy of projecting the current estimate
onto a hyperplane which makes the instantaneous error to be zero.
The difference is that HYPASS operates the projection in a functional
space (i.e., in a RKHS) while
KNLMS operates the projection in a Euclidean space of the coefficient
vector (see\cite{takizawa_icassp14,takizawa_tsp14}).
The multikernel adaptive filtering algorithms presented in
\cite{yukawa_sip10_multi,yukawa_eusipco11,yukawa_tsp12} are basically
extensions of the KNLMS algorithm.
Our recent study, on the other hand, reveals significant advantages of
HYPASS over KNLMS (cf.~\cite{hypass,takizawa_icassp13,takizawa_tsp14}).
It is therefore of significant interests how the two different
streams (multikernel adaptive filtering and HYPASS) meet.

\begin{figure}[t!]
 \psfrag{MKNLMS}[Bl][Bl][0.9]{MKNLMS \cite{yukawa_sip10_multi,yukawa_eusipco11,yukawa_tsp12}}
 \psfrag{CHYPASS}[Bl][Bl][0.9]{{\color{red}CHYPASS (Proposed)}}
 \psfrag{KNLMS}[Bl][Bl][0.9]{KNLMS \cite{richard09}}
 \psfrag{NORMA}[Bl][Bl][0.9]{NORMA \cite{kivinen04}}
 \psfrag{QKLMS}[Bl][Bl][0.9]{QKLMS \cite{chen_TNNLS12}}
 \psfrag{Dodd}[Bl][Bl][0.9]{Dodd {\it et al.}~\cite{dodd03}}
 \psfrag{HYPASS}[Bl][Bl][0.9]{HYPASS \cite{hypass,takizawa_tsp14}}
 \psfrag{Cartesian}[Bl][Bl][0.9]{{\color{red}Cartesian product}}
 \psfrag{formulation}[Bl][Bl][0.9]{{\color{red}formulation}}
\centering
 \includegraphics[width=10cm]{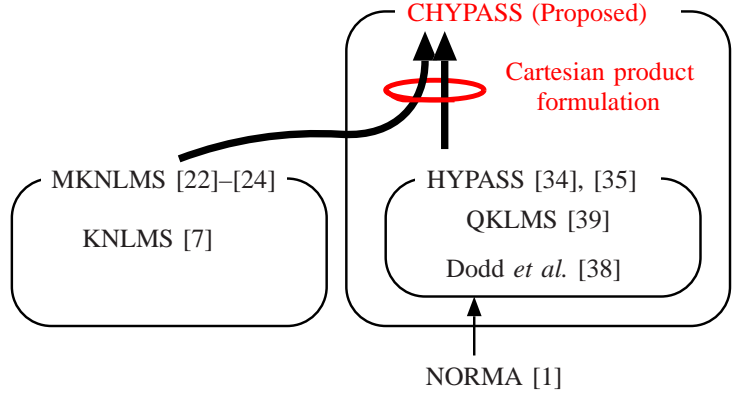}
 \caption{The orientation of the present study.
The two streams, MKNLMS (multikernel adaptive filtering) and HYPASS, are
 united into a single scheme (CHYPASS) based on the Cartesian-product formulation.
}\vspace{0em}
\label{fig:summary}
\end{figure}


In the present article,
we propose an efficient multikernel adaptive filtering algorithm
based on iterative orthogonal projections
in a functional space, inheriting the spirit of HYPASS
(see Fig.~\ref{fig:summary}).
A multikernel adaptive filter is characterized as 
a superposition of vectors lying in multiple RKHSs, namely
as a vector in the {\it sum space} of multiple RKHSs.
In general, a vector in the sum space can be decomposed,
in infinitely many ways, into vectors in the multiple RKHSs, and this would
cause a difficulty in computing the inner product in the sum space.
To avoid the difficulty, 
we first consider the particular case that any pair of the multiple
RKHSs intersects only trivially; 
i.e., any pair of the RKHSs shares only the zero vector.
It covers the important case of 
using linear and Gaussian kernels simultaneously
(see Corollary \ref{corollary:poly_gauss} in Section
\ref{subsec:sum_examples}).
In this case, the decomposition is unique,
which means that the sum space is the {\it direct sum} of the RKHSs,
and the inner product can be computed
easily in the sum space.
This allows us to derive an efficient algorithm 
by reformulating the HYPASS algorithm
in the sum space which is known to be a RKHS
(Theorem \ref{theorem:rk_sumspace}).
Due to the uniqueness of decomposition,
the sum space is isomorphic, as a Hilbert space,
to the {\it Cartesian-product} of the multiple RKHSs.
This implies that the same derivation is possible through 
the Cartesian formulation
instead of the sum-space formulation.
This is the key to extending the algorithm to the general case.

Now, let us turn our attention to
another important case of using multiple Gaussian kernels simultaneously.
It is widely known that Gaussian RKHSs have a nested structure
\cite{vert06, steinwart06, a_tanaka11} (see also Theorem
\ref{theorem:Gaussians} in Section \ref{subsec:general_examples}).
This means that the multiple-Gaussian case
is {\it not} covered by the first particular case.
We therefore consider the general case in which some pair of the RKHSs may
intersect non-trivially; i.e., some pair of the RKHSs may share common nonzero vectors.
In this case, the inner product in the sum space has no closed-form
expression, and hence it is generally intractable to derive an algorithm
through the sum-space formulation.
The inner product in the {\it Cartesian product}, on the other hand, is always expressed
in a closed form.
As a result, the algorithm formulated in the product space for the general case
boils down to the same formula 
as obtained from the sum-space algorithm for the first case.
The proposed algorithm is an iterative projection method in the Cartesian product
and, only in the first particular case, it can be viewed as a sum-space projection method.
The proposed algorithm is thus referred to as the Cartesian HYPASS (CHYPASS) algorithm.
The computational complexity is low due to 
a selective updating technique, which is also employed in HYPASS.
Numerical examples with toy models demonstrate that
(i) CHYPASS with linear and Gaussian kernels is effective 
in the case that 
the unknown system contains linear and nonlinear components and
(ii) CHYPASS with two Gaussian kernels is effective 
in the case that 
the unknown system contains high- and low- frequency components.
We also apply CHYPASS to real-world data and show its efficacy
over the KNLMS and HYPASS algorithms.

The rest of the paper is organized as follows.
Section \ref{sec:sum_space_model} presents the sum space model.
In Section \ref{sec:special_case}, we derive
the proposed algorithm through the sum-space formulation
for the particular case mentioned above.
We show that
the use of linear and single-Gaussian kernels corresponds to 
the particular case based on a theorem proved recently by Minh \cite{minh10}.
In Section \ref{sec:general_case}, we present the CHYPASS algorithm
for the general case as well as its computational complexity
for the two useful cases: the linear-Gaussian and two-Gaussian cases.
Section \ref{sec:numerical} presents numerical examples, followed by
concluding remarks in Section \ref{sec:conclusion}.

\section{Sum Space Model}
\label{sec:sum_space_model}

\subsection{Basic Mathematics}
\label{subsec:basic}

We denote by $\real$ and $\Natural$ the sets of all real numbers
and nonnegative integers, respectively.
Vectors and matrices are denoted by lower-case and upper-case
letters in bold-face, respectively.
The identity matrix is denoted by $\signal{I}$ and
the transposition of a vector/matrix is denoted by $(\cdot)^{\sf T}$.
We denote the null (zero) function by $0$.

Let $\mathcal{U}\subset \real^L$ and $\real$ be the input
and output spaces, respectively.
We consider a problem of estimating/tracking
a nonlinear unknown function $\psi:\mathcal{U}\rightarrow \real$
by means of sequentially arriving input-output measurements.
Our particular attention is focused 
on the case where $\psi$ contains several distinctive components;
e.g., linear and nonlinear (but smooth) components, high- and low-
frequency components, etc.
To generate a minimal model to describe 
such a multicomponent function $\psi$,
it would be natural to use multiple RKHSs
$(\euclidspace_1,\innerprod{\cdot}{\cdot}_{\euclidspace_1})$, 
$(\euclidspace_2,\innerprod{\cdot}{\cdot}_{\euclidspace_2})$, 
 $\cdots$,
$(\euclidspace_Q,\innerprod{\cdot}{\cdot}_{\euclidspace_Q})$ over
$\mathcal{U}$;
i.e., each of the $\euclidspace_q$s consists of functions mapping from
$\mathcal{U}$ to $\real$.
Here, $Q$ is the number of components of $\psi$ and
each RKHS is associated with each component.
The positive definite kernel associated with
the $q$th RKHS $\euclidspace_q$, $q\in\mathcal{Q}:=\{1,2,\cdots,Q\}$,
is denoted by
$\kappa_q:\mathcal{U}\times \mathcal{U}\rightarrow \real$,
and the norm induced by $\innerprod{\cdot}{\cdot}_{\euclidspace_q}$
is denote by $\norm{\cdot}_{\euclidspace_q}$.
The $\psi$ is modeled as an element of the sum space
\begin{equation*}
\euclidspace^+:=   \euclidspace_1 +
\euclidspace_2+\cdots+\euclidspace_Q
:=  \ \left\{\sum_{q\in\mathcal{Q}}
f_q:f_q\in\euclidspace_q\right\}.
\end{equation*}
Given an $f\in \euclidspace^+$, decomposition
$f=\sum_{q\in\mathcal{Q}}f_q$, $f_q\in\euclidspace_q$, is not
necessarily unique in general.
If such decomposition is unique for any $f\in \euclidspace^+$,
the sum space is specially called the {\it direct sum} of
$\euclidspace_q$s \cite{luenberger} and is usually indicated as
$\euclidspace^{+}=\euclidspace_1\oplus\euclidspace_2\oplus\cdots
\oplus\euclidspace_Q$.

\begin{theorem}[Reproducing kernel of sum space $\euclidspace^+$
 \cite{aronszajn50}]
\label{theorem:rk_sumspace}
The sum space $\euclidspace^+$ equipped with the norm
\begin{align}
\hspace*{-1.7em} \norm{f}_{\euclidspace^+}^2:=
\min\left\{\sum_{q\in\mathcal{Q}} \norm{f_q}^2_{\euclidspace_q} \mid
f=\sum_{q\in\mathcal{Q}} f_q,~
f_q\in\euclidspace_q
 \right\}, ~f\in\euclidspace^+,&
 \label{eq:norm_H}
\end{align}
is a RKHS with the reproducing kernel
$\kappa:=\sum_{q\in\mathcal{Q}} \kappa_q$.
\end{theorem}
{\it Proof:} One can apply \cite[Theorem in Part I Section 6]{aronszajn50} recursively
to verify the claim.\migip

\begin{theorem}
\label{theorem:rkhs_scolor}
 Let $\kappa:\mathcal{U}\times\mathcal{U}\rightarrow\real$ be
the reproducing kernel
of a real Hilbert space
$(\euclidspace,\innerprod{\cdot}{\cdot}_{\euclidspace})$.
Then, given an arbitrary $w>0$,
$\kappa_w(\signal{u},\signal{v})
:= w \kappa(\signal{u},\signal{v})$,
$\signal{u},\signal{v}\in\mathcal{U}$,
is the reproducing kernel of the RKHS
$(\euclidspace,\innerprod{\cdot}{\cdot}_{\euclidspace,w})$
with the inner product $\innerprod{\signal{u}}{\signal{v}}_{\euclidspace,w}:=
w^{-1} \innerprod{\signal{u}}{\signal{v}}_{\euclidspace}$,
 $\signal{u},\signal{v}\in\mathcal{U}$.
\end{theorem}
{\it Proof:} It is clear that
$\kappa_w(\cdot,\signal{u})\in\euclidspace$
for any $\signal{u}\in\mathcal{U}$.
Also, for any $f\in\euclidspace$ and $\signal{u}\in\mathcal{U}$,
we have
$\innerprod{f}{\kappa_w(\cdot,\signal{u})}_{\euclidspace,w}
=w^{-1}\innerprod{f}{w\kappa(\cdot,\signal{u})}_{\euclidspace}=f(\signal{u})$.
\migip

By Theorems \ref{theorem:rk_sumspace} and \ref{theorem:rkhs_scolor},
we can immediately obtain the following result.

\begin{corollary}[Weighted norm and reproducing kernel]
\label{corollary:weighted_norm}
Given any $w_q>0, ~q\in\mathcal{Q}$,
$\kappa_{\signal{w}}(\signal{u},\signal{v}):=\sum_{q\in\mathcal{Q}}$
$ w_q\kappa_q(\signal{u},\signal{v})$,
$\signal{u},\signal{v}\in\mathcal{U}$,
is the reproducing kernel of the sum space $\euclidspace^+$ 
equipped with the weighted norm $\norm{\cdot}_{\euclidspace^+,\signal{w}}$
defined as
$\norm{f}_{\euclidspace^+,\signal{w}}^2:=
\min\big\{\sum_{q\in\mathcal{Q}} w_q^{-1}\norm{f_q}^2_{\euclidspace_q} \mid
f=\sum_{q\in\mathcal{Q}} f_q,~
f_q\in\euclidspace_q \big\}$,
$f\in\euclidspace^+$.
\end{corollary}
Without loss of generality, we let $w_q=1$, $\forall q\in\mathcal{Q}$,
in the following.
For some batch processing techniques such as the kernel ridge regression,
the sum space $\euclidspace^+$ is easy to handle;
see Appendix \ref{appendix:krr_sum}.
For online/adaptive processing, on the other hand, it is hard
due to the fact that the inner product in $\euclidspace^+$
has no closed-form expression in general.
Fortunately, however, the inner product has a simple closed-form
expression in the case of {\it direct sum},
allowing us to build an adaptive algorithm in $\euclidspace^{+}$
as shown in Section \ref{sec:special_case}.

\subsection{Multikernel Adaptive Filter}

We denote by $\mathcal{D}_{q,n}\subset\{\kappa_q(\cdot,\signal{u})\mid
\signal{u}\in\mathcal{U}\}$
the {\it dictionary} constructed for the $q$th kernel at time $n\in\Natural$.
The {\it kernel-by-kernel dictionary subspaces}
are defined as
$\mathcal{M}_{q,n}:=
{\rm span} ~\mathcal{D}_{q,n}\subset\euclidspace_q$, $q\in\mathcal{Q}$,
$n\in\Natural$,
and their sum 
$\mathcal{M}_{n}^+:=\mathcal{M}_{1,n}+\mathcal{M}_{2,n}+\cdots +
\mathcal{M}_{Q,n}$ is the {\it dictionary subspace of the sum space $\euclidspace^+$}.
The multikernel adaptive filter at time $n$ is given in the following form:
\begin{equation}
 \varphi_n:=\sum_{q\in\mathcal{Q}} \varphi_{q,n}\in
\mathcal{M}_{n-1}^+\subset\euclidspace^{+}, \ n\in\Natural,
\end{equation}
where $\varphi_{q,n}\in\mathcal{M}_{q,n-1}$.
Thus, the dictionary $\mathcal{D}_{q,n}$ contains the atoms (vectors)
that form the next estimate $\varphi_{q,n+1}$.
If some a priori information is available, we may
accordingly define an initial dictionary $\mathcal{M}_{q,-1}$ and
an initial filter $\varphi_0$.
Otherwise, we simply let $\mathcal{M}_{q,-1}:=\{0\}$
and $\varphi_0:=0$.
We assume that 
\sinq{active} elements in $\mathcal{D}_{q,n-1}$ remain
in $\mathcal{D}_{q,n}$ so that
\begin{equation}
 \varphi_n\in \mathcal{M}_n^+\cap \mathcal{M}_{n-1}^+.
\end{equation}

\section{Special Case: $\euclidspace_p\cap \euclidspace_q=\{0\}$ 
for any $p\neq q$}
\label{sec:special_case}

In this section, we focus on the particular case that 
$\euclidspace_p\cap \euclidspace_q=\{0\}$ 
for any $p\neq q$.
This is the case of direct sum
(in which
any $f\in\euclidspace^+$ can be {\it decomposed uniquely} into
$f= \sum_{q\in\mathcal{Q}}f_q$,
$f_q\in\euclidspace_q$) and includes some useful examples
as will be discussed precisely in Section \ref{subsec:sum_examples}.
Due to the unique decomposability,
the norm in \refeq{eq:norm_H} is reduced to
\begin{equation}
 \norm{f}_{\euclidspace^{+}}^2 =
\sum_{q\in\mathcal{Q}} \norm{f_q}^2_{\euclidspace_q},
 \label{eq:norm_H_direct_sum}
\end{equation}
and accordingly the inner product
between 
$f= \sum_{q\in\mathcal{Q}}f_q\in \euclidspace^{+}$ and
$g= \sum_{q\in\mathcal{Q}}g_q\in \euclidspace^{+}$
is given by
\begin{equation}
 \innerprod{f}{g}_{\euclidspace^{+}}:=
\sum_{q\in\mathcal{Q}} \innerprod{f_q}{g_q}_{\euclidspace_q}.
 \label{eq:innerprod_H_direct_sum}
\end{equation}
It is clear that, under the correspondence
between $f$ and the $Q$-tuple
$(f_q)_{q\in\mathcal{Q}}$,
the sum space $\euclidspace^{+}$ is isomorphic
to the Cartesian product
\begin{align*}
\euclidspace^{\times}:=  \euclidspace_1 \times
\euclidspace_2 \times \cdots\times \euclidspace_Q
:=  \ \left\{(f_1,f_2,\cdots,f_Q) :f_q\in\euclidspace_q,
~q\in\mathcal{Q}\right\},
\end{align*}
which is a real Hilbert space equipped
with the inner product defined as
\begin{align}
 \innerprod{f}{g}_{\euclidspace^{\times}}:=\sum_{q\in\mathcal{Q}}
\innerprod{f_q}{g_q}_{\euclidspace_q}, ~ f=(f_q)_{q\in\mathcal{Q}}, ~
g=(g_q)_{q\in\mathcal{Q}}\in\euclidspace^{\times}.
\end{align}

\subsection{Examples}
\label{subsec:sum_examples}

We present three cerebrated examples of positive definite kernel below
(see, e.g., \cite{schoelkopf01}).
\begin{example}[Positive definite kernels]
~
 \begin{enumerate}
  \item Linear kernel: Given $c\geq 0$,
 \begin{equation}
  \kappa_{\rm L}(\signal{x},\signal{y}):=\signal{x}^{\sf T}\signal{y} +c,~
\signal{x},\signal{y}\in\mathcal{U}.
\label{eq:linear_kernel}
 \end{equation}
  \item Polynomial kernel: Given $c\geq 0$ and
	$m\in\Natural^*:=\Natural\setminus\{0\}$,
 \begin{equation}
  \kappa_{\rm P}(\signal{x},\signal{y}):=(\signal{x}^{\sf T}\signal{y}+c)^m,~
\signal{x},\signal{y}\in\mathcal{U}.
\label{eq:polynomial_kernel}
 \end{equation}
  \item Gaussian kernel (normalized): Given $\sigma>0$,
 \begin{equation}
  \kappa_{{\rm G},\sigma}(\signal{x},\signal{y}):=
\frac{1}{(\sqrt{2\pi}\sigma)^L}
\exp\left(-\frac{\norm{\signal{x}-\signal{y}}_{\real^L}^2}{2\sigma^2}\right), ~
\signal{x},\signal{y}\in\mathcal{U}.
\label{eq:Gaussian_kernel}
 \end{equation}
 \end{enumerate}
\end{example}

For the linear kernel, $c=1$ is a typical choice.
If one knows that the linear component of $\psi$ is zero-passing,
one can simply let $c=0$.
The following theorem has been shown by Minh in 2010 \cite{minh10}.

\begin{theorem}[\cite{minh10}]
\label{theorem:minh10}
Let $\mathcal{U}\subset\real^L$ be any set with nonempty interior
and $\euclidspace_{\kappa_{{\rm  G},\sigma}}$
the RKHS associated with
a Gaussian kernel  $\kappa_{{\rm  G},\sigma}(\signal{x},\signal{y})$ 
for an arbitrary $\sigma>0$ together with the input space $\mathcal{U}$.
Then, $\euclidspace_{\kappa_{{\rm  G},\sigma}}$
does not contain any polynomial on $\mathcal{U}$, including the nonzero
constant function.
\end{theorem}

The following corollary is obtained as 
a direct consequence of Theorem \ref{theorem:minh10}.

\begin{corollary}[Polynomial and Gaussian RKHSs]
\label{corollary:poly_gauss}
Assume that the input space $\mathcal{U}$ has nonempty interior.
Given arbitrary $c\geq 0$, $m\in\Natural^*$, and $\sigma>0$,
denote by $\euclidspace_{\kappa_{\rm P}}$ and
$\euclidspace_{\kappa_{{\rm G},\sigma}}$ the RKHSs associated
respectively with the polynomial and Gaussian kernels
$\kappa_{\rm P}$ and $\kappa_{{\rm G},\sigma}$.
Then,
\begin{equation}
\euclidspace_{\kappa_{\rm P}}\cap \euclidspace_{\kappa_{{\rm
 G},\sigma}}=\{0\}.
\label{eq:polynomial_gaussian}
\end{equation}
In particular, \refeq{eq:polynomial_gaussian} for $m=1$
 implies that
\begin{equation}
\euclidspace_{\kappa_{\rm L}}\cap 
\euclidspace_{\kappa_{{\rm G},\sigma}}=\{0\}.
\label{eq:linear_gaussian}
\end{equation}
\end{corollary}

We mention that a (manually-tuned) convex combination of linear and Gaussian kernels
has been used in \cite{gilcacho13} 
{\it within a single-kernel adaptive filtering framework}
for nonlinear acoustic echo cancellation.
The case of linear plus Gaussian kernels is of particular interest
when the unknown function $\psi$ contains linear and nonlinear (smooth) components
\cite{haerdle00,espinoza05,xu_ac09}.
(Our recent work in \cite{yukawa_icassp15} is devoted to this important case.)
We will present a dictionary design for this case in the following subsection.

\subsection{Dictionary Design: Linear Plus Gaussian Case}
\label{subsec:sum_dictionary}

The dictionaries are designed on a kernel-by-kernel basis.
With Corollary \ref{corollary:poly_gauss} in mind,
we present a possible dictionary design
for the case of $Q=2$ with $\kappa_1:=\kappa_{\rm L}$ for $c:=1$ and
$\kappa_2:=\kappa_{{\rm G},\sigma}$,
assuming that the input space $\mathcal{U}$ has nonempty interior.
Due to the interior assumption on $\mathcal{U}$,
it is seen that the dimension of $\euclidspace_{1}$ is $L+1$.
It is clear that
$\kappa_{1}(\cdot,\signal{0})=c$ and
$\kappa_{1}(\cdot,\signal{e}_j) - \kappa_{1}(\cdot,\signal{0})  =
\signal{e}_j^{\sf T}(\cdot)$, 
where $\signal{e}_j\in\real^L$ is the unit vector having one at the
$j$th entry and zeros elsewhere.
Based on this observation,
one can see that 
\begin{equation}
 \mathcal{D}_{{1}}:=\{\kappa_{1}(\cdot,\signal{e}_j)
-\kappa_{1}(\cdot,\signal{0}) \}_{j=1}^L
\cup \{\kappa_{1}(\cdot,\signal{0}) \}
\end{equation}
gives an orthonormal basis of the $L+1$ dimensional space $\euclidspace_{1}$.
We thus let $\mathcal{D}_{1,n}:=\mathcal{D}_{1}$ for all $n\in\Natural$,
which implies that 
$\mathcal{M}_{1,n}=\euclidspace_1$ and hence
$P_{\mathcal{M}_{1,n}}(\kappa_1(\cdot,\signal{u}))=\kappa_1(\cdot,\signal{u})$
for any $\signal{u}\in\mathcal{U}$.
Note that, in the case of $c:=0$, 
the dimension of $\euclidspace_{1}$ is $L$
and one can remove $\kappa_{1}(\cdot,\signal{0})$ from the
dictionary $\mathcal{D}_1$.

On the other hand, the dictionary $\mathcal{D}_{2,n}$ for the Gaussian kernel
needs to be constructed in online fashion.
In general, one may consider growing and pruning strategies to
construct an adequate dictionary.
A growing strategy is given as follows:
(i) start with $\mathcal{D}_{2,-1}:=\emptyset$, and
(ii) add a new candidate $\kappa_2(\cdot,\signal{u}_n)$ into the dictionary
at each time $n\in\Natural$ only when it is sufficiently novel.
In this case,
$\mathcal{D}_{2,n}=\{\kappa_2(\cdot,\signal{u}_j)\}_{j\in\mathcal{J}_n}$
for some $\mathcal{J}_n\subset\{0,1,2,\cdots,n\}$.
As a possible novelty criterion for the present example,
we use Platt's criterion \cite{platt91} with a slight modification:
$\kappa_2(\cdot,\signal{u}_n)$ is regarded to be novel if
$c_{\euclidspace_2}(\mathcal{D}_{2,n},\kappa_2(\cdot,\signal{u}_n))
=\displaystyle\max_{j\in\mathcal{J}_n}
\exp\left(-
\dfrac{\norm{\signal{u}_j-\signal{u}_n}_{\real^L}^2}{2\sigma^2}
\right)
<\delta$
for some $\delta\in(0,1)$ and if
$\abs{d_n - \varphi_n(\signal{u}_n)}^2>\varepsilon\abs{\varphi_n(\signal{u}_n)}^2$ 
for some $\varepsilon>0$.
Here, given a RKHS $\euclidspace$ with its associated kernel
$\kappa:\mathcal{U}\times\mathcal{U}\rightarrow \real$
and a dictionary $\mathcal{D}:=\{\kappa(\cdot,\signal{u}_j)\}_{j\in\mathcal{J}}$
with an index set $\mathcal{J}\subset\Natural$,
the coherence is defined as
$c_{\euclidspace}(\mathcal{D},\kappa(\cdot,\signal{u})):=
\displaystyle\max_{j\in\mathcal{J}}
\dfrac{\abs{\kappa(\signal{u}_j,\signal{u})}}
{\sqrt{\kappa(\signal{u}_j,\signal{u}_j)}
\sqrt{\kappa(\signal{u},\signal{u})}}$.
Pruning can be done based, e.g., on $\ell_1$ regularization;
see, e.g., \cite{yukawa_tsp12,chen_IJCCN12,gao14,takizawa_icassp14}.

\subsection{Adaptive Learning Algorithm in Sum Space}
\label{subsec:sum_space_algorithm}

At every time instant $n\in\Natural$,
a new measurement $\signal{u}_n$ and $d_n$ arrives, and
$\varphi_n$ is updated to $\varphi_{n+1}$ based on the new measurement.
A question is how to exploit the new measurement
for obtaining a better estimator within the subspace $\mathcal{M}_n^+$.
A simple strategy accepted widely in adaptive filtering is the way of 
the normalized least mean square (NLMS) algorithm \cite{nagumo,albert67}, projecting the
current estimate onto a zero-instantaneous-error hyperplane
in a relaxed sense.
See \cite{yagu_paper,yuya_jasp09,thslya_IEEESPMAG11} and the references therein
for more about the projection-based adaptive methods.
As we assume that the search space is restricted to $\mathcal{M}_n^+$,
we consider the following hyperplane in $\mathcal{M}_n^+$:
\begin{equation}
 \Pi_n:=\left\{f\in\mathcal{M}_{n}^{+}\mid 
f(\signal{u}_n)=\innerprod{f}{\kappa(\cdot,\signal{u}_n)}_{\euclidspace^{+}}=d_n\right\}.
\label{eq:Pi_def}
\end{equation}
Note here that $\Pi_n$ can also be represented as 
$$
\Pi_n=\mathcal{M}_n^+\cap \Pi_{n}^{\mathcal{H}^+},
$$
where
$ \Pi_{n}^{\mathcal{H}^+}:=\left\{f\in\mathcal{H}^+
\mid f(\signal{u}_n)=\innerprod{f}{\kappa(\cdot,\signal{u}_n)}_{\euclidspace^{+}}=d_n\right\}$
is a hyperplane in the whole space $\euclidspace^{+}$.
The update equation is given by
\begin{equation}
 \varphi_{n+1}:=\varphi_n + \lambda_n (P_{\Pi_n} (\varphi_n) - \varphi_n)\in
  \mathcal{M}_{n}^{+},
~ n\in\Natural,
\label{eq:update_equation_sum_space}
\end{equation}
where $\lambda_n\in(0,2)$ is the step size.
Here, for any $f\in\euclidspace^+$ and any linear variety (affine set) $\mathcal{V}\subset\euclidspace^+$,
$P_{\mathcal{V}} (f):=\argmin_{g\in\mathcal{V}}\norm{f-g}_{\euclidspace^{+}}$
denotes the orthogonal projection of $f$ onto the set $\mathcal{V}$ \cite{luenberger}.
The projection $P_{\Pi_n}(\varphi_n)$ in \refeq{eq:update_equation_sum_space}
can be computed with the following theorem.

\begin{theorem}[Orthogonal projection in sum space]
\label{prop:projection}
Let $(\euclidspace_q,\innerprod{\cdot}{\cdot}_{\euclidspace_q})$, 
$q\in\mathcal{Q}$, be a RKHS over $\mathcal{U}$ with its reproducing
 kernel $\kappa_q$ and
define the sum space $\euclidspace^+:=   \euclidspace_1 +
\euclidspace_2+\cdots+\euclidspace_Q$ with its kernel
$\kappa:=\sum_{q\in\mathcal{Q}} \kappa_q$.
Let $\mathcal{M}_q$ be a subspace of $\euclidspace_q$ and
define its sum
$\mathcal{M}^{+}:=
\mathcal{M}_1+ \mathcal{M}_2+\cdots +\mathcal{M}_Q$.
Also define
$\Pi:=\left\{f\in\mathcal{M}^+\mid
       f(\signal{u})=\innerprod{f}{\kappa(\cdot,\signal{u})}_{\euclidspace^{+}}\right.$
$\left.=d\right\}$
for some $\signal{u}\in\mathcal{U}$ and $d\in\real$.
Then, the following hold.
\begin{enumerate}
  \item For any $\phi\in\mathcal{M}^{+}$,
\begin{equation}
 P_{\Pi} (\phi)=\phi + \frac{d -
  \phi(\signal{u})}{\norm{P_{\mathcal{M}^{+}}(\kappa(\cdot,\signal{u}))
  }_{\euclidspace^{+}}^2} P_{\mathcal{M}^{+}}(\kappa(\cdot,\signal{u})).
\label{eq:P_Pi_phi}
\end{equation}
  \item Assume that
$\euclidspace_p\cap \euclidspace_q=\{0\}$ 
for any $p\neq q$.
Then, for any $f=\sum_{q\in\mathcal{Q}}f_q\in\euclidspace^{+}$
with $(f_q)_{q\in\mathcal{Q}}\in\euclidspace^\times$,
\begin{equation}
P_{\mathcal{M}^{+}}(f)=\sum_{q\in\mathcal{Q}}
 P_{\mathcal{M}_q}(f_q).
\label{eq:P_M_kappa_u}
\end{equation}
 \end{enumerate}
 
\end{theorem}
{\it Proof:} See \cite{takizawa_tsp14} for \refeq{eq:P_Pi_phi}.
Define $\mathcal{M}^\times:=\mathcal{M}_1\times\mathcal{M}_2\times\cdots\times\mathcal{M}_Q\subset\euclidspace^\times$.
By \refeq{eq:norm_H_direct_sum}, we have
\begin{align*}
P_{\mathcal{M}^{+}}(f) &=\argmin_{g\in \mathcal{M}^{+}}
 \norm{f - g}_{\euclidspace^{+}} \\
&=\argmin_{\begin{tabular}{c}
{\footnotesize $\sum_{q\in\mathcal{Q}}g_q\in\mathcal{M}^{+}$}\\
{\footnotesize s.t.~$(g_q)_{q\in\mathcal{Q}}\in\mathcal{M}^\times$}
	   \end{tabular}}
\sum_{q\in\mathcal{Q}} \norm{f_q -
 g_q}_{\euclidspace_q}^2 \\
&=\sum_{q\in\mathcal{Q}} 
\argmin_{g_q\in \mathcal{M}_q}
\norm{f_q - g_q}_{\euclidspace_q} \\
&=\sum_{q\in\mathcal{Q}} P_{\mathcal{M}_q}(f_q).
\end{align*}
\migip

We stress that Theorem \ref{prop:projection}.2 only holds
under the assumption that
$\euclidspace_p\cap \euclidspace_q=\{0\}$ 
for any $p\neq q$.
From \refeq{eq:P_Pi_phi} and \refeq{eq:P_M_kappa_u}, 
the computation of $P_{\Pi}(\phi)$ involves
$P_{\mathcal{M}_q}(\kappa_q(\cdot,\signal{u}))$ which can be computed
with the following lemma.

\begin{lemma}[\cite{luenberger}]
\label{lemma:PMf}
Let $\euclidspace$ denote a RKHS associated
 with an input space $\mathcal{U}$ and a positive definite kernel
 $\kappa:\mathcal{U}\times \mathcal{U}\rightarrow\real$.
Let $\mathcal{D}:=\{\kappa(\cdot,\signal{x}_j)\}_{j=1}^r$ for
 $\signal{x}_j\in\mathcal{U}$, $j=1,2,\cdots, r$, and
$\mathcal{M}:={\rm span}~\mathcal{D}$.
Then, given any $f\in\mathcal{H}$, 
\begin{equation}
 P_{\mathcal{M}}(f)=\sum_{j=1}^{r} \alpha_j
  \kappa(\cdot,\signal{x}_j), ~ \alpha_j\in\real,
\end{equation}
where the coefficient vector
 $\signal{\alpha}:=[\alpha_1,\alpha_2,\cdots,\alpha_r]^{\sf T}\in\real^r$ 
is characterized as a solution of the following normal equation:
\begin{equation}
 \signal{K}\signal{\alpha}=\signal{b},
\end{equation}
 where  $\signal{K}\in\real^{r\times r}$ is the kernel (or Gram) matrix whose
$(i,j)$ entry is $\kappa(\signal{x}_i,\signal{x}_j)$ and
$\signal{b}:=[f(\signal{x}_1),
f(\signal{x}_2),
\cdots,$
$f(\signal{x}_r) ]^{\sf T}\in\real^r$.
\end{lemma}
If $f=\kappa(\cdot,\signal{x}_j)$ for some
$j\in\{1,2,\cdots,r\}$, we obtain a trivial solution $\alpha_j=1$
and $\alpha_i=0$ for $i\neq j$ which yields
$P_{\mathcal{M}}(\kappa(\cdot,\signal{x}_j))=\kappa(\cdot,\signal{x}_j)$.

\subsection{The Sum-space HYPASS Algorithm:
Complexity Issue and Practical Remedy}
\label{subsec:sum_complexity}

Theorem \ref{prop:projection} and Lemma \ref{lemma:PMf}
indicate that the computation of $P_{\Pi_n}(\varphi_n)$ in
\refeq{eq:update_equation_sum_space} would involve
the inversion of the kernel matrix (if invertible) for each kernel
as well as the multiplication of the inverse matrix by a vector,
where the size of the kernel matrix and the vector is determined by the
dictionary size.
Note here that this computation is unnecessary when
the dictionary is orthonormal such as in the case of linear kernel
(see Section \ref{subsec:sum_examples}).
In the case of Gaussian kernels, the inversion needs to be computed 
and a practical remedy to reduce the complexity
is the selective update which is described below.

Let
$\tilde{\mathcal{D}}_{q,n}$ be
a selected subset of the dictionary $\mathcal{D}_{q,n}$
for the $q$th kernel $\kappa_q$.
For instance, 
in the case of $\kappa_1:=\kappa_{\rm L}$ and $\kappa_2:=\kappa_{{\rm G},\sigma}$
(the case of linear and Gaussian kernels),
one can simply let 
$\tilde{\mathcal{D}}_{1,n}:=\mathcal{D}_1$
and design $\tilde{\mathcal{D}}_{2,n}$
by selecting a few $\kappa_2(\cdot,\signal{u}_j)$s in $\mathcal{D}_{2,n}$
that are most coherent to $\kappa_2(\cdot,\signal{u})$;
i.e., choose $\kappa_2(\cdot,\signal{u}_j)$ such that
$\kappa_2(\signal{u}_j,\signal{u})$ is the largest
\cite{hypass,takizawa_icassp13,takizawa_tsp14}.
In other words, we choose $\signal{u}_j$s
such that $\norm{\signal{u}_j-\signal{u}_n}_{\real^L}$ is the smallest
(or the neighbors of $\signal{u}_n$ are collected in short).
Geometrically, the maximal coherence implies the least angle 
between $\kappa_2(\cdot,\signal{u}_j)$ and
$P_{\mathcal{M}_{2,n}}(\kappa_2(\cdot,\signal{u}_n))$
which gives the direction of update in the exact form of
\refeq{eq:update_equation_sum_space}; see
\cite{takizawa_icassp13,takizawa_tsp14}.
This means that the selected $\kappa_2(\cdot,\signal{u}_j)$
approximates the exact direction $P_{\mathcal{M}_{2,n}}(\kappa_2(\cdot,\signal{u}_n))$
best in the Gaussian dictionary $\mathcal{D}_{2,n}$.
The coherence-based selection is therefore reasonable,
as justified by numerical examples in Section \ref{sec:numerical}.

Now, we define the subspace spanned by each selected dictionary as
\begin{equation}
\tilde{\mathcal{M}}_{q,n}:={\rm span}~
\tilde{\mathcal{D}}_{q,n}
\subseteq \mathcal{M}_{q,n}
\end{equation}
and its sum 
$\tilde{\mathcal{M}}_n^{+}:=\tilde{\mathcal{M}}_{1,n}\oplus
\tilde{\mathcal{M}}_{2,n}\oplus\cdots\oplus\tilde{\mathcal{M}}_{Q,n}$.
To update only the coefficient(s) of the selected dictionary element(s)
and keep the other coefficients fixed, the next estimate $\varphi_{n+1}$
is restricted to
$\mathcal{V}_n^+:=\tilde{\mathcal{M}}_n^{+} + \varphi_n$,
rather than to $\mathcal{M}_n^+$ (cf.~\refeq{eq:Pi_def}).
Accordingly, the update equation in \refeq{eq:update_equation_sum_space} is
modified into
\begin{equation}
 \varphi_{n+1}:=\varphi_n + 
\lambda_n (P_{\tilde{\Pi}_n} (\varphi_n) - \varphi_n)\in
  \mathcal{M}_{n}^{+},
~ n\in\Natural,
\label{eq:sum_space_hypass}
\end{equation}
where
\begin{equation}
 \tilde{\Pi}_n:=
\left\{f\in\mathcal{V}_n^+\mid f(\signal{u}_n)=
\innerprod{f}{\kappa(\cdot,\signal{u}_n)}_{\euclidspace^{+}}=d_n\right\}.
\end{equation}
In the trivial case that
$\tilde{\mathcal{D}}_{q,n} = \mathcal{D}_{q,n}$  for all $q\in\mathcal{Q}$,
\refeq{eq:sum_space_hypass} is reduced to
\refeq{eq:update_equation_sum_space}.
Indeed, the algorithm in \refeq{eq:sum_space_hypass} is a {\it sum-space extension
of the HYPASS algorithm} proposed in \cite{hypass}.
The following proposition can be used,
together with
Theorem \ref{prop:projection} and Lemma \ref{lemma:PMf},
 to compute $P_{\tilde{\Pi}_n} (\varphi_n)$.

\begin{proposition}
\label{prop:projection_mod}

For any $\phi\in\euclidspace^+$ and a subspace
$\mathcal{M}^+$ of $\euclidspace^+$, let
 $\mathcal{V}^+ :=\mathcal{M}^{+} + \phi$
and $\Pi_{\mathcal{V}^+}:=\left\{f\in\mathcal{V}^+\mid
       f(\signal{u})=\right.$
$\left.\innerprod{f}{\kappa(\cdot,\signal{u})}_{\euclidspace^{+}}=d\right\}$
for some $\signal{u}\in\mathcal{U}$ and $d\in\real$.
Then,  for any $f\in\mathcal{V}^+$,
\begin{equation}
 P_{\Pi_{\mathcal{V}^+}} (f)=f + \frac{d -
  f(\signal{u})}{\norm{P_{\mathcal{M}^{+}}(\kappa(\cdot,\signal{u}))
  }_{\euclidspace^{+}}^2} P_{\mathcal{M}^{+}}(\kappa(\cdot,\signal{u})).
\label{eq:P_Pi_tilde_phi}
\end{equation}

\end{proposition}
The computational complexity of the proposed algorithm
under the selective updating strategy stated above
will be given in Section \ref{subsec:complexity}.

\section{General Case}
\label{sec:general_case}

We consider the general case in which
it may happen that
$\euclidspace_p\cap \euclidspace_q\neq \{0\}$
for some $p\neq q$.
In this case, 
given an $f\in\euclidspace^+$,
decomposition
$f= \sum_{q\in\mathcal{Q}}f_q$,
$f_q\in\euclidspace_q$, is {\it not necessarily unique},
and thus Theorem \ref{prop:projection}.2 does not generally hold anymore,
although Theorem \ref{prop:projection}.1 and 
Proposition \ref{prop:projection_mod} still hold.
This implies that
$P_{\tilde{\Pi}_n}(\varphi_n)$ in \refeq{eq:P_Pi_tilde_phi}
cannot be obtained simply in general.
In the following, we show that this issue can be overcome by considering
the Cartesian product $\euclidspace^{\times}$
rather than sticking to the sum space $\euclidspace^{+}$.

\subsection{Examples}
\label{subsec:general_examples}

We show below, in a slightly general form, the known fact that
the class of Gaussian kernels has a nested structure.
\begin{theorem}
\label{theorem:Gaussians}
Let $\mathcal{U}\subset\real^L$ be an arbitrary subset and
$\kappa_1:=w_1\kappa_{{\rm G},\sigma_1}$ and
$\kappa_2:=w_2\kappa_{{\rm G},\sigma_2}$ Gaussian kernels for
 $\sigma_1>\sigma_2>0$ and $w_1,w_2>0$.
Then, the associated RKHSs $\euclidspace_1$ and $\euclidspace_2$
satisfy the following.
\begin{enumerate}
 \item $ \euclidspace_1 \subset  \euclidspace_2$.
 \item $\sqrt{w_1} \norm{f}_{\euclidspace_1} \geq 
\sqrt{w_2} \norm{f}_{\euclidspace_2}$
for any $f\in \euclidspace_1$.
\end{enumerate}
\end{theorem}
{\it Proof:}
Let $\kappa(\signal{u},\signal{v}):=\kappa_{{\rm
G},\sigma_2}(\signal{u},\signal{v}) - \kappa_{{\rm
G},\sigma_1}(\signal{u},\signal{v})$,
$\signal{u},\signal{v}\in\real^L$
 and define
$\gamma:\real^L\rightarrow\real,~\signal{u}\mapsto\kappa(\signal{u},\signal{0})$.
Then, its Fourier transform
is given by $\hat{\gamma}(\signal{w}):=
\int_{\real^L} \gamma(\signal{u})\exp\left(-\sqrt{-1}\signal{u}^{\sf T}\signal{w}\right)d\signal{u}=
\exp\left(-\frac{\sigma_2^2}{2}\norm{\signal{w}}_{\real^L}^2\right)
- \exp\left(-\frac{\sigma_1^2}{2}\norm{\signal{w}}_{\real^L}^2\right)$
for $\signal{w}\in\real^L$.
The function $\hat{\gamma}(\signal{w})$ is clearly bounded
and also satisfies $\hat{\gamma}(\signal{w})\geq 0$ because $\sigma_1>\sigma_2>0$.
Hence, Bochner's theorem \cite{berlinet04} ensures that
$\gamma(\signal{u}-\signal{v})=\kappa(\signal{u},\signal{v})$ is a positive
definite kernel on $\real^L$, and so on $\mathcal{U}$ as well by the
definition of positive definite kernels.
Applying \cite[Theorem I in Part I Section 7]{aronszajn50},
we obtain $\euclidspace_{\kappa_{{\rm G},\sigma_1}}\subset 
\euclidspace_{\kappa_{{\rm G},\sigma_2}}$
and $\norm{f}_{\euclidspace_{\kappa_{{\rm G},\sigma_1}}} \geq 
\norm{f}_{\euclidspace_{\kappa_{{\rm G},\sigma_2}}}$,
which verifies the case of $w_1=w_2=1$.
This is generalized to any $w_1,w_2>0$
because one can verify 
under the light of Theorem \ref{theorem:rkhs_scolor} that
$\sqrt{w_q} \norm{f}_{\euclidspace_q} = 
\norm{f}_{\euclidspace_{\kappa_{{\rm G},\sigma_q}}}$
for any $f\in \euclidspace_q = \euclidspace_{\kappa_{{\rm G},\sigma_q}}$
($q=1,2$).
We remark here that the two RKHSs $\euclidspace_q$ 
(associated with $\kappa_q:=w_q\kappa_{{\rm G},\sigma_q}$)
and
$\euclidspace_{\kappa_{{\rm G},\sigma_q}}$
(associated with $\kappa_{{\rm G},\sigma_q}$)
shares the common elements ---
this is what is meant by $\euclidspace_q = \euclidspace_{\kappa_{{\rm
G},\sigma_q}}$ above ---
but are equipped with different inner products when $w_q\neq 1$.
\migip

There exist several articles that show some results related to 
Theorem \ref{theorem:Gaussians}.
For instance, a special case of Theorem \ref{theorem:Gaussians} for
$\mathcal{U}=\real^L$ and $w_1=w_2=1$ can be found in \cite{vert06}.
The proof in  \cite{vert06} is based on a characterization of a Gaussian RKHS
in terms of Fourier transform.
It is straightforward to generalize it to any subset
$\mathcal{U}\subset\real^L$ with nonempty interior by exploiting
\cite[Theorem 1]{minh10} 
which gives another characterization of a Gaussian RKHS.
Note that Theorem \ref{theorem:Gaussians} holds with no assumption
on the existence of interior of $\mathcal{U}$.
To verify Theorem \ref{theorem:Gaussians},
one can also follow the way in \cite{a_tanaka11} which proves
the case of $L=1$ and $w_1=w_2=1$ by using another theorem
in place of Bochner's theorem.
The inclusion operator ``id'' appearing in \cite{steinwart06}
would imply Theorem \ref{theorem:Gaussians}.1, and
a result related to a special case of Theorem \ref{theorem:Gaussians}.2
for $w_1=w_2=1$ can also be found in \cite[Corollary 6]{steinwart06}.

\subsection{Dictionary Design: Two Gaussian Case}
\label{subsec:general_examples}

We present our dictionary selection strategy
for the case of two Gaussian kernels
$\kappa_1:=w_1\kappa_{{\rm G},\sigma_1}$ and
$\kappa_2:=w_2\kappa_{{\rm G},\sigma_2}$  for
$\sigma_1>\sigma_2>0$ and $w_1,w_2>0$.
In analogy with Section \ref{subsec:sum_examples}, we define the
dictionary for each kernel as
$\mathcal{D}_{q,n}:=\{\kappa_q(\cdot,\signal{u}_j)\}_{j\in\mathcal{J}_{q,n}}$
for $q=1,2$, 
where $\mathcal{J}_{q,n}\subset\{0,1,2,\cdots,n\}$.
For the kernel $\kappa_1$, we simply adopt the coherence criterion
\cite{richard09}:
$\kappa_1(\cdot,\signal{u}_n)$ is regarded to be novel if
$c_{\euclidspace_1}(\mathcal{D}_{1,n},\kappa_1(\cdot,\signal{u}_n))
<\delta_1$
for some $\delta_1\in(0,1)$.
The kernel $\kappa_2$ is complementary in the sense that
it only needs to be used
in those regions (of the input space $\mathcal{U}$) where
the unknown system $\psi$ contains
high frequency components which make the \sinq{wider} kernel $\kappa_1$ underfit the system.
To do so,
a new element $\kappa_2(\cdot,\signal{u}_n)$ enters into the
dictionary $\mathcal{D}_{2,n}$
only when all of the following three conditions are satisfied:
(i) $\kappa_1(\cdot,\signal{u}_n)$ 
does not enter into the dictionary $\mathcal{D}_{1,n}$
(the no-simultaneous-entrance condition),
(ii) $c_{\euclidspace_2}(\mathcal{D}_{2,n},\kappa_2(\cdot,\signal{u}_n))
<\delta$
for some $\delta\in(0,1)$ (the small-coherence condition), and
(iii) $\abs{d_n - \varphi_n(\signal{u}_n)}^2 > \varepsilon
\abs{\varphi_n(\signal{u}_n)}^2$ for some
$\varepsilon>0$ (the large-error condition).

\subsection{The Cartesian HYPASS Algorithm}
\label{subsec:general_case_algorithm}

By virtue of the isomorphism between 
the sum space $\euclidspace^+$ and 
the product space $\euclidspace^\times$
in the case of $\euclidspace_p\cap \euclidspace_q=\{0\}$,
$\forall p\neq q$,
the arguments as in Section \ref{sec:special_case}
can be translated into the product space $\euclidspace^\times$.
(See \cite{yukawa_ieice_TechRep14} for the direct derivation in the
product space.)
Fortunately, the translated arguments can be applied to
the general case, including the case that
$\euclidspace_p\cap \euclidspace_q\neq \{0\}$
for some $p\neq q$.
This is because, even when $f\in\euclidspace^+$ can be decomposed
in two different ways like $f=\sum_{q\in\mathcal{Q}}f_q = 
\sum_{q\in\mathcal{Q}}\hat{f}_q$, the two functions
$\sum_{q\in\mathcal{Q}}f_q$ and
$\sum_{q\in\mathcal{Q}}\hat{f}_q$
are distinguished in the product space
as $(f_q)_{q\in\mathcal{Q}}\neq
(\hat{f}_q)_{q\in\mathcal{Q}}\in\euclidspace^\times$.
Therefore, the product-space formulation
delivers the following algorithm for the general case:
\begin{align}
 \varphi_{n+1} \!:=\!  \varphi_n \!\!+\!\!\lambda_n 
\frac{d_n-\varphi_n(\signal{u}_n)}
{\! \displaystyle\sum_{q\in\mathcal{Q}}
\!\bigg\|P_{\tilde{\mathcal{M}}_{q,n}}\!\!
(\kappa_q(\cdot,\signal{u}_n)\!)\bigg\|_{\euclidspace_q}^2}
\!\! \displaystyle\sum_{q\in\mathcal{Q}}
\! P_{\tilde{\mathcal{M}}_{q,n}}\!\!(\kappa_q(\cdot,\signal{u}_n)\!), ~ n\in\Natural,
\label{eq:product_space_hypass}
\end{align}
 which is seemingly identical to \refeq{eq:sum_space_hypass}
under Proposition \ref{prop:projection_mod}.
We emphasize here that
\refeq{eq:sum_space_hypass} can be written in the form of
\refeq{eq:product_space_hypass}
only in the case of $\euclidspace_p\cap \euclidspace_q=\{0\}$,
$\forall p\neq q$.
Namely, 
in the case of $\euclidspace_p\cap \euclidspace_q\neq\{0\}$,
$\exists p\neq q$,
\refeq{eq:product_space_hypass} can be regarded
as a hyperplane projection algorithm in the product space
$\euclidspace^\times$, but not in the sum space $\euclidspace^+$.
We call the general algorithm in \refeq{eq:product_space_hypass}
the {\it Cartesian HYPASS (CHYPASS)} algorithm, 
since it is a product-space extension of the HYPASS algorithm.
In the case of two Gaussian kernels,
the coherence-based selective updating strategy discussed
in Section \ref{subsec:sum_complexity}
is applied to each Gaussian kernel.

\subsection{Alternative Algorithm: Parameter-space Approach}
\label{subsec:alternatives}

We present a simple alternative to the CHYPASS algorithm.
Let us parametrize $\varphi_{q,n}$ by
\begin{equation}
 \varphi_{q,n}=\sum_{f\in\mathcal{D}_{q,n}}
h_{f,n}^{(q)} f, ~q\in\mathcal{Q}, ~n\in\Natural,
\end{equation}
where $h_{f,n}^{(q)}\in\real$.
Then, $\varphi_{q,n}(\signal{u}_n)$ can be expressed as
\begin{equation}
 \varphi_{q,n}(\signal{u}_n) =
\sum_{f\in\mathcal{D}_{q,n}}h_{f,n}^{(q)} f(\signal{u}_n) =
\signal{h}_{q,n}^{\sf T} \signal{k}_{q,n}
\end{equation}
by defining the vectors 
$\signal{h}_{q,n}\in\real^{r_{q,n}}$ and
$\signal{k}_{q,n}\in\real^{r_{q,n}}$ appropriately that consist 
of $h_{f,n}^{(q)}$s and $f(\signal{u}_n)$s for $f\in\mathcal{D}_{q,n}$,
respectively, where $r_{q,n}:=\abs{\mathcal{D}_{q,n}}$.
Concatenating $Q$ vectors yields
$\signal{h}_n:=[\signal{h}_{1,n}^{\sf T}
\signal{h}_{2,n}^{\sf T}
\cdots
\signal{h}_{Q,n}^{\sf T}
 ]^{\sf T}\in\real^{r_n}$
and 
$\signal{k}_n:=[\signal{k}_{1,n}^{\sf T}
\signal{k}_{2,n}^{\sf T}
\cdots
\signal{k}_{Q,n}^{\sf T}
 ]^{\sf T}\in\real^{r_n}$
with $r_n:=\sum_{q\in\mathcal{Q}} r_{q,n}$.
Then, $\varphi_n(\signal{u}_n)$ is simply expressed by
\begin{equation}
 \varphi_n(\signal{u}_n)=\signal{h}_n^{\sf T}\signal{k}_n
=\innerprod{\signal{h}_n}{\signal{k}_n}_{\real^{r_n}}.
\end{equation}
One can therefore build an algorithm that projects
the current coefficient vector $\signal{h}_n$
onto the following zero-instantaneous-error hyperplane in the Euclidean
space:
\begin{equation}
 H_n:=\left\{
\signal{h}\in\real^{r_n}\mid
\innerprod{\signal{h}}{\signal{k}_n}_{\real^{r_n}}=d_n
\right\}.
\end{equation}
This is the idea of the alternative algorithm.
The next coefficient vector $\hat{\signal{h}}_{n+1}\in\real^{r_n}$
containing $h_{f,n+1}^{(q)}$s for $f\in\mathcal{D}_{q,n}$ is computed as
\begin{equation}
 \hat{\signal{h}}_{n+1}:=
\signal{h}_n + \lambda_n \left(P_{H_n}(\signal{h}_n)-\signal{h}_n\right),~n\in\Natural,
\end{equation}
where $\lambda_n\in(0,2)$.
At the next iteration, if $\mathcal{D}_{q,n}= \mathcal{D}_{q,n+1}$
($\Rightarrow r_n=r_{n+1}$),
$\signal{h}_{n+1}\in\real^{r_{n+1}}$ is given by
$\hat{\signal{h}}_{n+1}$ itself.
Otherwise, $\signal{h}_{n+1}$
is obtained with $\hat{\signal{h}}_{n+1}$ and
$h_{f,n+1}^{(q)}=0$ for
$f\in\mathcal{D}_{q,n+1}\setminus\mathcal{D}_{q,n}$.
We call the alternative algorithm the multikernel NLMS (MKNLMS)
since it is essentially the same as
the algorithm presented in \cite[Section III.A]{yukawa_tsp12}
except that the dictionary is designed individually for each kernel.
MKNLMS with two Gaussian kernels with individual dictionaries
has been studied earlier in \cite{ishida13}.

\subsection{Computational Complexity}
\label{subsec:complexity}

The computational complexity is discussed
in terms of the number of
multiplications required for each update, including
the dictionary update, 
for CHYPASS and MKNLMS in the linear-Gaussian and two-Gaussian cases, respectively.
The complexity is summarized in Table \ref{tb:complexity}.

\subsubsection{Linear-Gaussian case}
\label{subsubsec:complexity_linear_Gaussian}

The complexity of CHYPASS is $(L+3)r_{2,n} + 3L + \min\{L,s_n\} + O(s_n^3)$,
where $r_{2,n}:=\abs{\mathcal{D}_{2,n}}$ is the size of the Gaussian
dictionary $\mathcal{D}_{2,n}$ and 
$s_n:=|\tilde{\mathcal{D}}_{2,n}|$ is 
the size of its selected subset $\tilde{\mathcal{D}}_{2,n}$.
Here, $\abs{S}$ denotes the cardinality of a set $S$.
The term $O(s_n^3)$ is for the inversion of an $s_n\times s_n$
submatrix (which is supposed to be small) of the $r_{2,n}\times r_{2,n}$ kernel
matrix.
If one does not make use of the selective updating strategy and
updates all the coefficients of $\mathcal{D}_{2,n}$,
the matrix inversion of the $r_{2,n}\times r_{2,n}$ kernel
matrix can be computed in the $O((r_{2,n}-1)^2)$ complexity
by using the formula for the inverse of a partitioned matrix 
together with the matrix inversion lemma \cite{horn_johnson85}.
In addition to that, the inversion needs to be computed only when the
dictionary is updated.
The complexity in this computationally demanding case is
$(L+5)r_{2,n} + 3L + \min\{L,r_{2,n}\} +r_{2,n}^2$.
The complexity of MKNLMS is
$(L+5)r_{2,n} + 3L +4 + \min\{L,r_{2,n}\}$.

\subsubsection{Two-Gaussian case}
\label{subsubsec:complexity_two_Gaussian}

Assume that, for both Gaussian kernels,
the number of coefficients updated at the $n$th iteration
is equal to $s_n$.
Let $r_n:=r_{1,n}+r_{2,n}$ with
$r_{q,n}:=\abs{\mathcal{D}_{q,n}}$ for $q=1,2$.
The complexity of CHYPASS in this case
is $(L+3)r_n + O(s_n^3)$.
In the computationally demanding case of no coefficient selection,
the complexity is $(L+5)r_n + r_{1,n}^2 + r_{2,n}^2 + \min\{r_{1,n}, r_{2,n}\}$
for the same reason as described in Section \ref{subsubsec:complexity_linear_Gaussian}.
The complexity of MKNLMS in this case is
$(L+5)r_n + \min\{r_{1,n},r_{2,n}\}$.

\begin{table}[t]
\caption{Computational complexity of the algorithms.
The number $s_n$ of selected coefficients typically satisfies
$s_n\leq 5$.}
\label{tb:complexity}
\vspace*{-1em} 
\begin{center}
\begin{tabular}{|c|c|}
\hline 
NLMS & $3L+2 $\\ \hline
KNLMS & $(L+5)r_n +2$ \\ \hline
HYPASS & $(L+3)r_n + O(s_n^3)$ \\ \hline
CHYPASS& $(L+3)r_{2,n} + 3L + \min\{L,s_n\} + O(s_n^3)$ \\ 
(Linear-Gaussian)& \\ \hline
CHYPASS&$(L+3)r_n + O(s_n^3)$ \\ 
(Two-Gaussian)& \\ \hline
MKNLMS & $(L+5)r_{2,n} + 3L + \min\{L,r_{2,n}\}+4$\\
(Linear-Gaussian)& \\ \hline
MKNLMS &$(L+5)r_n + \min\{r_{1,n},r_{2,n}\}+4 $\\
(Two-Gaussian)& \\ \hline
\end{tabular}\vspace*{-1em} 
\end{center}
\end{table}

\subsubsection{Efficiency of CHYPASS}
\label{subsubsec:efficiency}

In analogy with HYPASS \cite{hypass,takizawa_tsp14}
and KNLMS \cite{richard09},
the dictionary size of CHYPASS is finite
under the dictionary construction rules presented in Sections
\ref{sec:special_case} and \ref{sec:general_case},
provided that the input space
$\mathcal{U}$ is compact (cf.~\cite{richard09}).
This property comes directly from the fact that
the coherence is exploited in a part of the dictionary construction.
To enhance the efficiency of CHYPASS, one may extend
the shrinkage-based pruning strategy 
that has been proposed for HYPASS in \cite{takizawa_icassp14}.
To keep the dictionary size bounded strictly by a prespecified number,
one can extend the simple technique presented for MKNLMS in
\cite{yukawa_apsipa13}
as well as the pruning strategy.

We emphasize that CHYPASS (as well as MKNLMS)
has a potential to be more efficient
than the single kernel approaches such as HYPASS and KNLMS
whenever the unknown system $\psi$ contains multiple components.
This is because the use of multiple kernels allows to represent
such a \sinq{multi-component} function $\psi$ {\it with a smaller size of dictionary (i.e., more compactly)},
as shown in the following section.

\section{Numerical Examples}
\label{sec:numerical}

We show the efficacy of the proposed algorithm
for three toy examples and two real data.\footnote{
Another experimental result for a larger real dataset will be presented
in a conference \cite{yukawa_icassp15}.
}
Throughout the section, we present
the curves for CHYPASS and MKNLMS
with linear and Gaussian kernels 
in red and magenta colors, respectively,
and those for CHYPASS and MKNLMS
with two Gaussian kernels in green and light-green colors, respectively.
The curves for the existing single-kernel algorithms, KNLMS \cite{richard09} and
HYPASS \cite{hypass}, are presented
in blue and light-blue colors.
It is worth mentioning that,
in the particular case that $s_n=1$ and a Gaussian kernel is employed,
HYPASS is reduced to QKLMS \cite{chen_TNNLS12};
this is the case for Section \ref{subsec:toy}, but not for Section
\ref{subsec:real_data}.

\begin{table}[t]
\caption{Parameter settings and complexities for Experiment A1.}
\label{tb:A1}
\vspace*{-1em} 
\begin{center}
\begin{tabular}{|c|c|c|c|}
\hline 
&\multicolumn{2}{|c|}{parameter}&complexity\\ \hline
NLMS & \multicolumn{2}{|c|}{$\lambda_n=0.1$}  & 5\\\hline
KNLMS & $\lambda_n=0.1$ & $\delta=0.99$  & 193\\ \cline{1-1}\cline{3-4}
HYPASS & $\sigma=0.5$ &$\delta=0.99$, $s_n=1$ & 132\\ \cline{1-1}\cline{3-4}
MKNLMS &$\varepsilon=0.05$ &$\delta=0.95$, $w_2=0.5$ &  107 \\
(Linear-Gaussian) & & & \\
 \cline{1-1}\cline{3-4}
CHYPASS&  &  $\delta=0.95$, $w_2=0.5$ &75\\ 
(Linear-Gaussian)& &$s_n=1$ &\\ \hline
\end{tabular}\vspace*{-1em} 
\end{center}
\end{table}

\begin{figure}[t!]
\centering
\subfigure[]{
 \includegraphics[width=7.5cm]{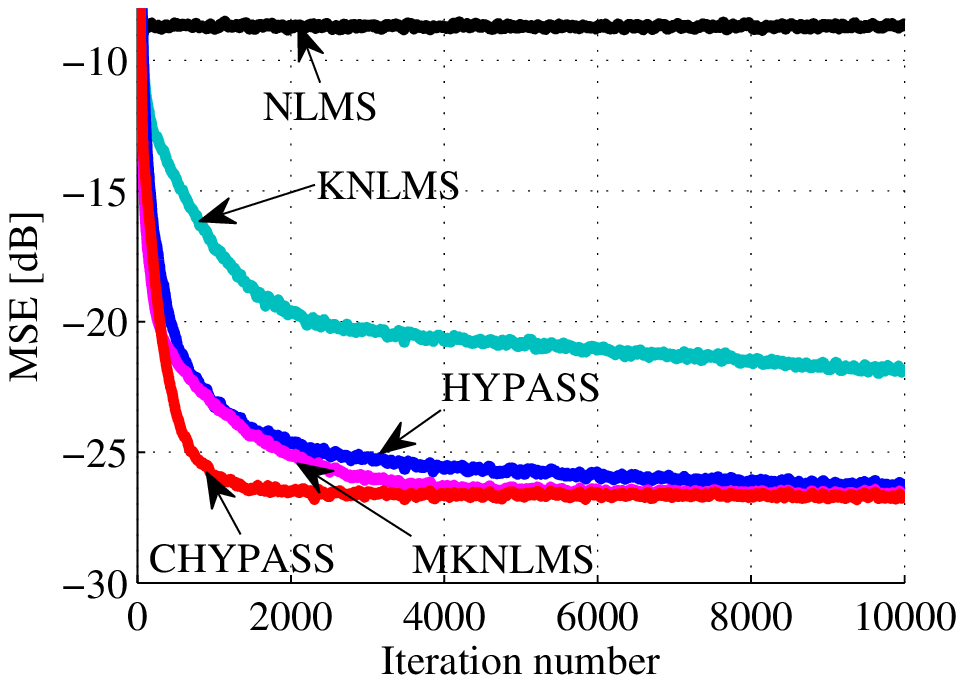}
}
\subfigure[]{
 \includegraphics[width=7.5cm]{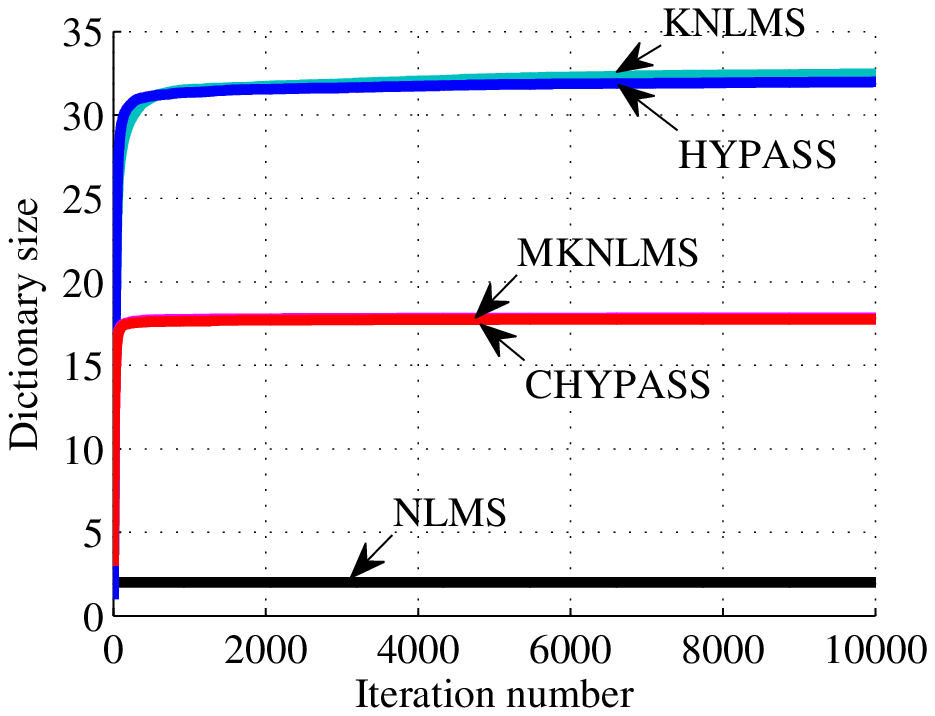}
}
\subfigure[]{
 \includegraphics[width=6cm]{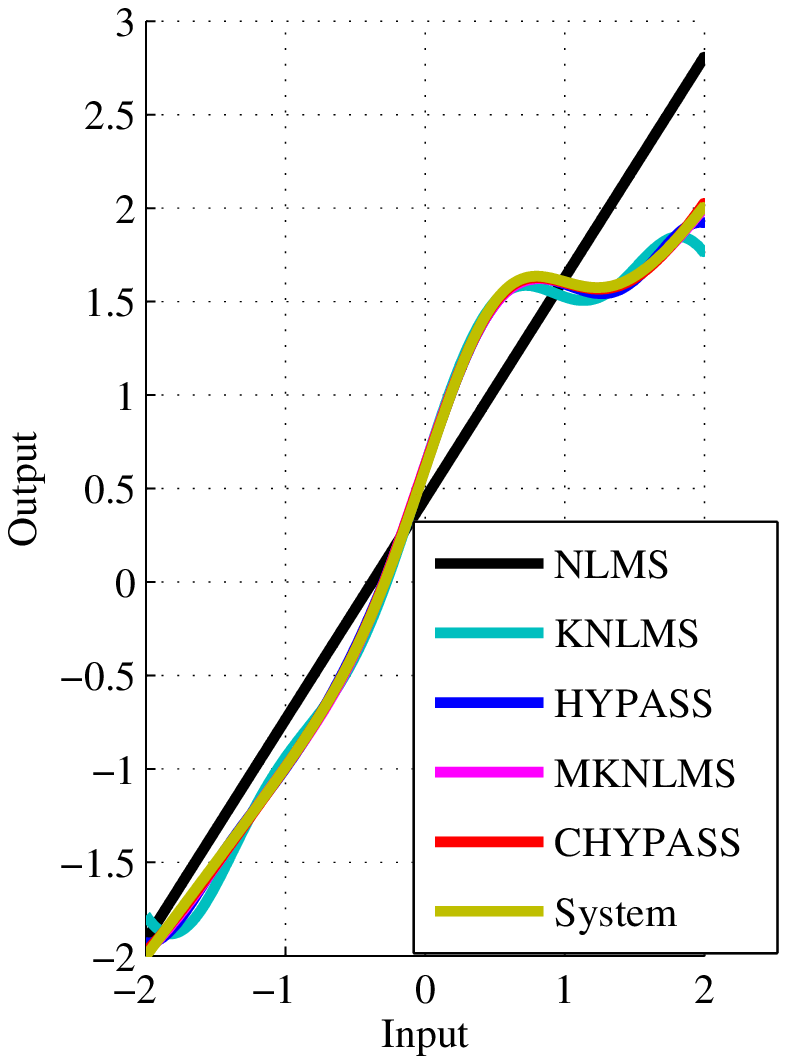}
 \includegraphics[width=6cm]{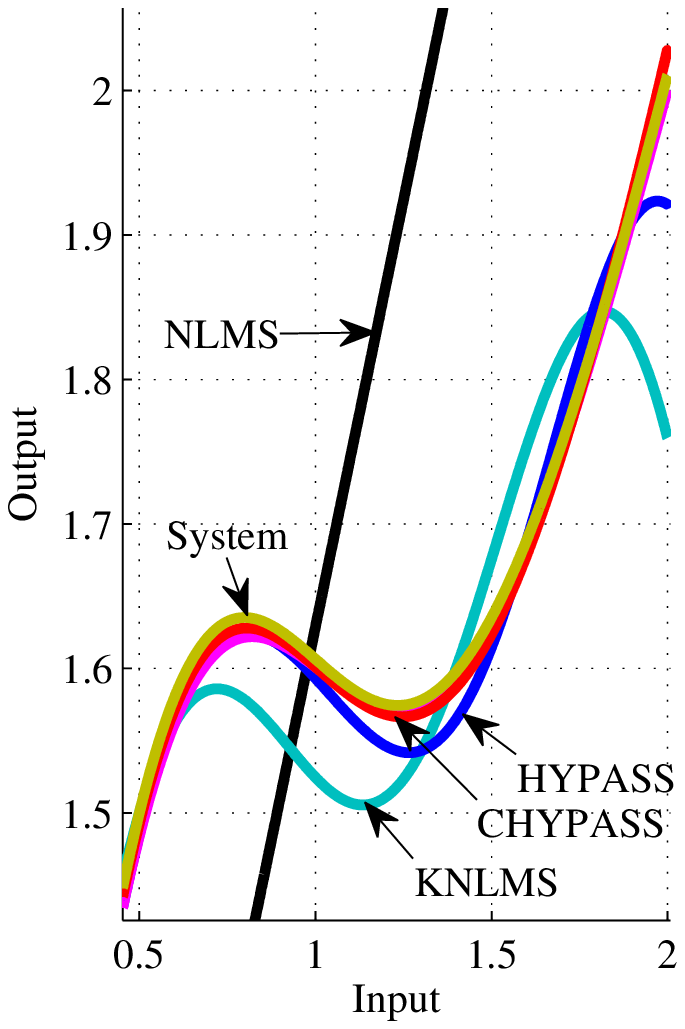}
}
 \caption{Results for Experiment A1: the linear-Gaussian case.}\vspace{0em}
\label{fig:LG}
\end{figure}

\subsection{Toy Models}
\label{subsec:toy}

\subsubsection{Experiment A1 - Linear Plus Gaussian Case}
\label{subsubsec:LG}

We consider the following linear-Gaussian model: $\psi(u):= u +
\exp\left(-\dfrac{(u-0.5)^2}{2\times 0.5^2}\right)$ for
$u\in\mathcal{U}:=(-2,2)\subset \real$ (i.e., $L=1$).
We compare the performance of the proposed multikernel adaptive
filtering algorithm (CHYPASS) with its alternative (MKNLMS)
as well as KNLMS and HYPASS.
For fairness, we adopt the same novelty criterion as described in
Section \ref{subsec:sum_examples}, which is basically Platt's criterion
\cite{platt91}, for KNLMS and HYPASS in all experiments.
For the multikernel adaptive filtering algorithms,
we employ the linear kernel $\kappa_1:=w_1\kappa_{\rm L}$ for $c:=1$ and
a Gaussian kernel $\kappa_2:=w_2\kappa_{{\rm G},\sigma}$
for $\sigma:=0.5$ (i.e., $Q=2$);
the weight is chosen as $w_1=w_2=0.5$
($w_1+w_2=1$).
An input sequence $(u_n)_{n\in\Natural}\subset \mathcal{U}$ is randomly
drawn from the uniform distribution over the input space $\mathcal{U}$.
The output of the unknown system is corrupted by 
an additive white Gaussian noise; the observed data is given by
$d_n:=\psi(u_n)+v_n$ 
with the Gaussian noise $v_n\sim \mathcal{N}(0,2.0\times10^{-3})$, $n\in\Natural$.
We test 300 independent trials and compute the mean squared error (MSE)
and the mean dictionary size
by averaging the values of
$(d_n-\varphi_n(u_n))^2$ and
$r_n:= r_{1,n} + r_{2,n}$, respectively,
over the 300 trials at each iteration $n\in\Natural$.
The parameters and complexities for each algorithm are summarized in Table \ref{tb:A1}.

Fig.~\ref{fig:LG} shows 
(a) the MSE learning curves, 
(b) the evolutions of the dictionary size, and
(c) an instance of the final estimate of each algorithm
as well as the system $\psi$ to be estimated.
For reference, the results of NLMS (the special case of
CHYPASS with $Q=1$, $\kappa_1:=\kappa_{\rm L}$ for $c=1$) are included.
The mean dictionary size was:
KNLMS 31.9, HYPASS 31.6, MKNLMS 17.8, and CHYPASS 17.7.
It is seen that both CHYPASS and MKNLMS outperform
their single-kernel counterparts with lower complexity.
This is due to the simultaneous use of linear and Gaussian kernels
under the multikernel adaptive filtering framework.
In the left panel of Fig.~\ref{fig:LG}(c), it is seen that 
all the nonlinear algorithms can estimate $\psi$ well globally.
The right panel shows the local behaviors
in the specific range $[0.5,2.0]$ of the input space.
One can see that the multikernel algorithms better estimate
$\psi$ than the single-kernel ones.

\subsubsection{Experiment A2 - Sinusoid Plus Gaussian Case}
\label{subsubsec:GG}

We consider the following model which has a low-frequency
component (sinusoid) and 
a high-frequency component (Gaussian):
$\psi(u):= \sin\left(\dfrac{\pi}{3}u\right) -
\exp\left(-\dfrac{(u-0.5)^2}{2\times 0.1^2}\right)$ for
$u\in\mathcal{U}:=(-2,2)\subset \real$ (i.e., $L=1$).
We compare CHYPASS and MKNLMS with
KNLMS and HYPASS.
For the multikernel adaptive filtering algorithms,
we employ two Gaussian kernels
 $\kappa_1 := w_1\kappa_{{\rm G},\sigma_1}$ and
 $\kappa_2 :=w_2\kappa_{{\rm G},\sigma_2}$
for $w_2 := 0.1$, $w_1:=1-w_2$,
$\sigma_1:=1.0$, and $\sigma_2:=0.02$ (i.e., $Q=2$).
The input and observed data sequences are
generated in a way similar to the previous experiment
with the noise variance $1.0\times 10^{-3}$.
The parameters and complexities for each algorithm are summarized in Table \ref{tb:A2}.

\begin{figure}[t!]
\centering
\subfigure[]{
 \includegraphics[width=7.5cm]{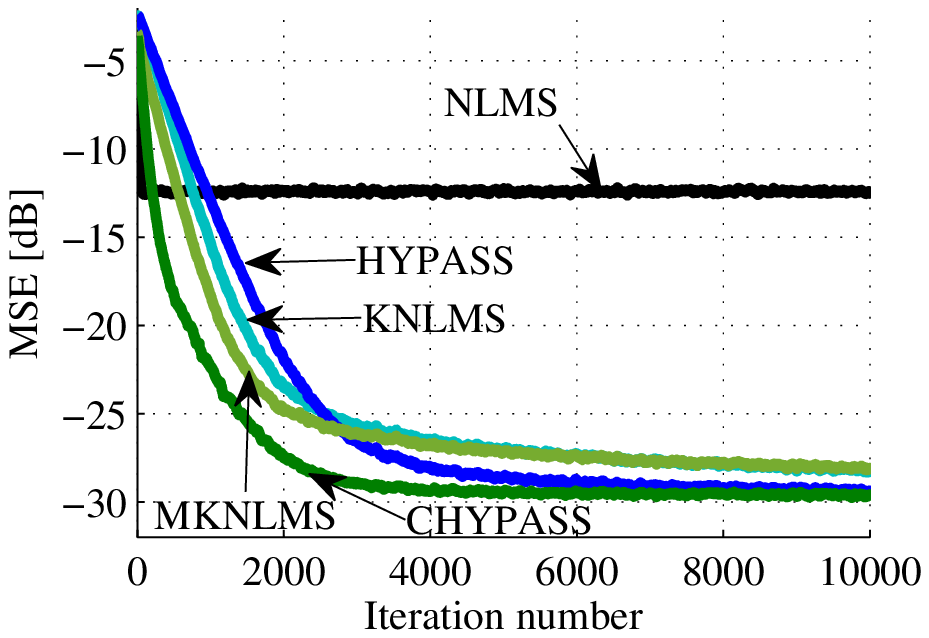}
}
\subfigure[]{
 \includegraphics[width=7.5cm]{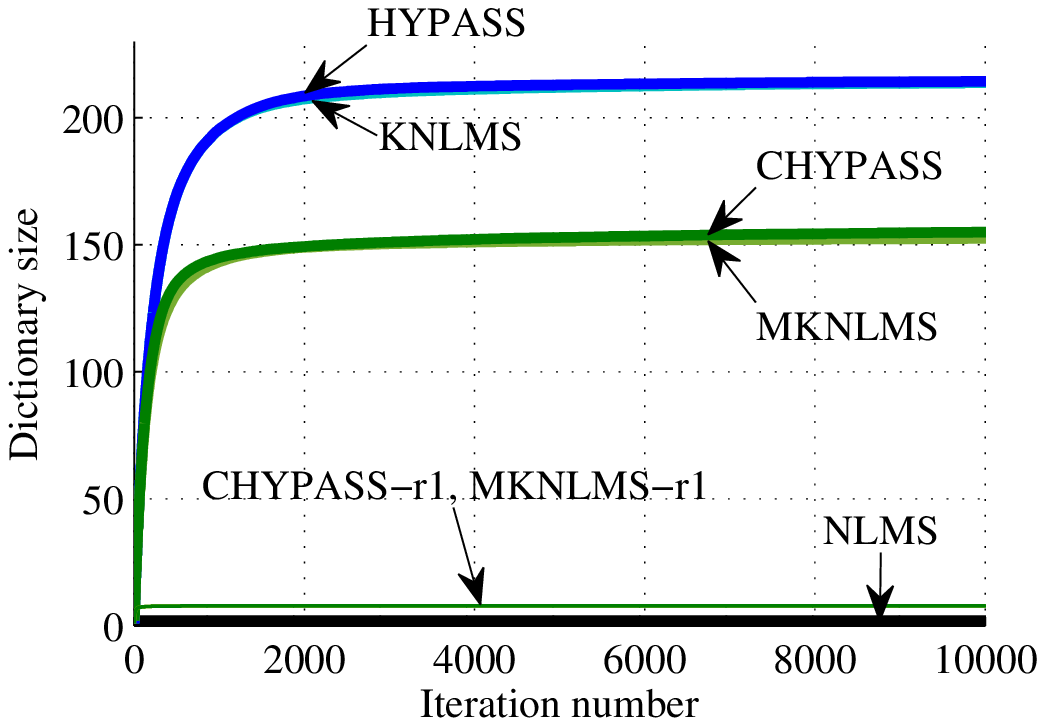}
}
\subfigure[]{
 \includegraphics[width=6cm]{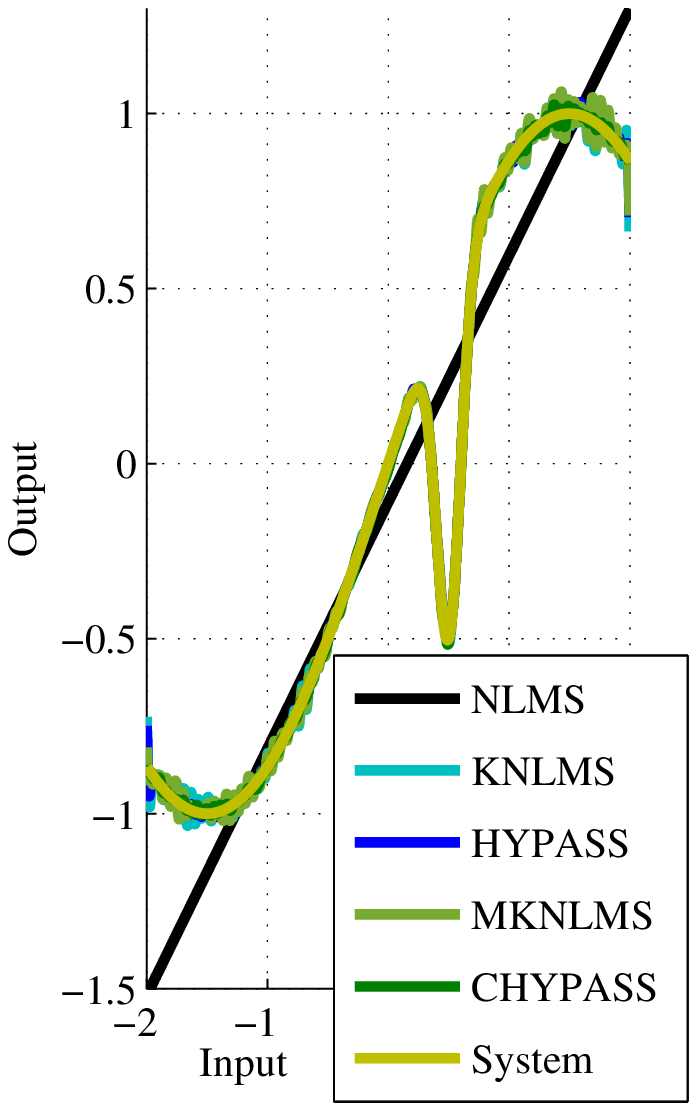}
 \includegraphics[width=6cm]{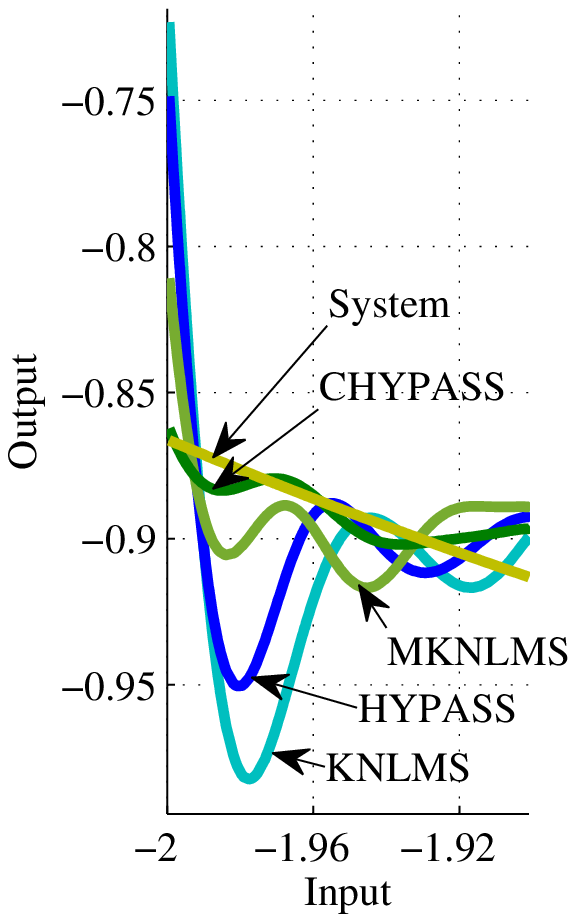}
}
 \caption{Results for Experiment A2: the sinusoid-Gaussian case.}\vspace{0em}
\label{fig:GG}
\end{figure}

Fig.~\ref{fig:GG} depicts the results.
In Fig.~\ref{fig:GG}(b),
the curves labeled as CHYPASS-r1 and MKNLMS-r1 show
the evolution of $r_{1,n}$ for each algorithm.
The mean dictionary size was: 
KNLMS 205.0, HYPASS 205.8, MKNLMS 147.6, and CHYPASS 149.3.
As in the results of Experiment A1,
CHYPASS and MKNLMS outperform their single-kernel counterparts with lower complexity.
It is also seen that CHYPASS significantly outperforms MKNLMS.
This is because the autocorrelation matrix of the kernelized input
vector $\signal{k}_n$ has a large eigenvalue spread, whereas
the condition number is improved in CHYPASS by using another metric
(cf.~\cite{takizawa_icassp14}).
Fig.~\ref{fig:GG}(c) shows that
the multikernel algorithms better estimate $\psi$ around the edge.

\subsubsection{Experiment A3 - Partially Linear Case}
\label{subsubsec:Partially_linear}

\begin{figure}[t!]
\centering
\subfigure[]{
 \includegraphics[width=7.5cm]{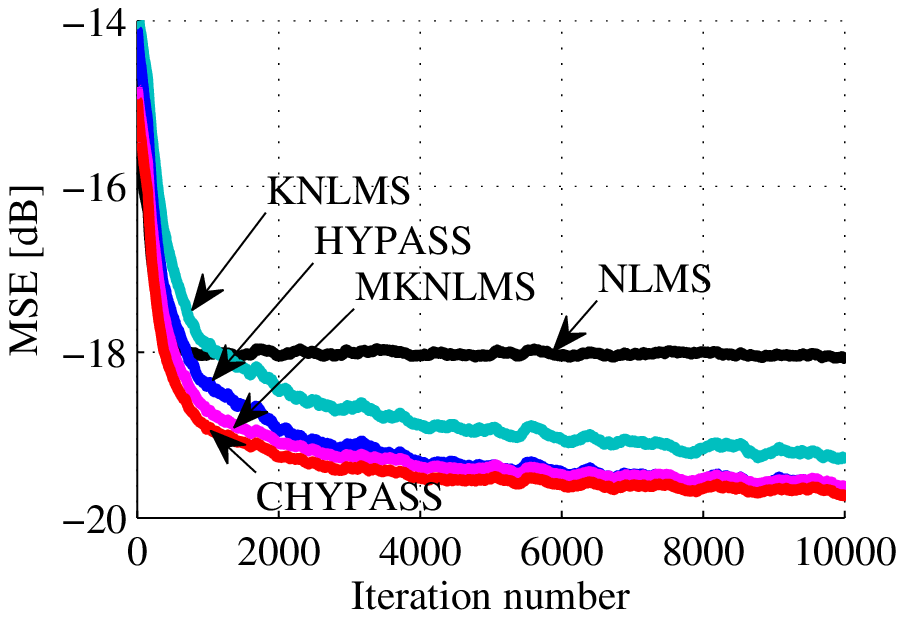}
}
\subfigure[]{
 \includegraphics[width=7.5cm]{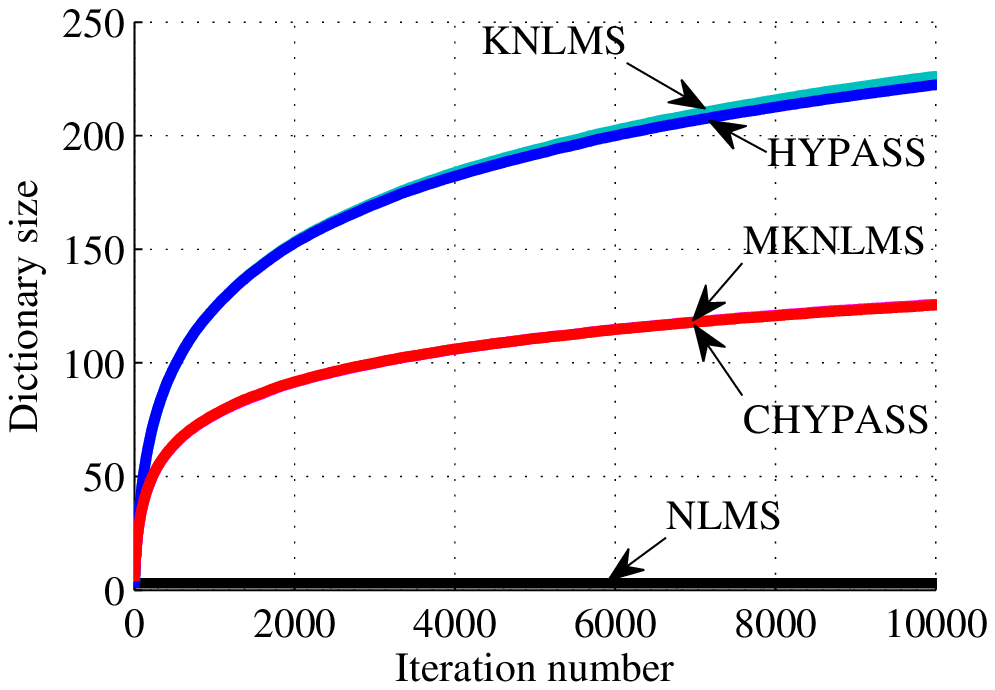}
}
 \caption{Results for Experiment A3: the partially linear case.}\vspace{0em}
\label{fig:Nonlinear_dynamics}
\end{figure}

We consider the following nonlinear dynamic system
which has a partially linear structure \cite{xu_ac09}:
$d_n:=0.5d_{n-1} +0.2 x_n +0.3 \sin (d_{n-1}x_n)
 + v_n$, $n\in\Natural$.
Here, $x_n\sim\mathcal{N}(0,1)$ is the excitation signal
and $v_n\sim\mathcal{N}(0,1.0\times 10^{-2})$ is the noise.
Each datum $d_n$ is a function of $u_n$ and $ d_{n-1}$
and is therefore predicted with
$\signal{u}_n:=[u_n, d_{n-1}]^{\sf T}$
(i.e., $L=2$).
We employ the linear kernel $\kappa_1:=w_1\kappa_{\rm L}$ for $c:=1$ and
a Gaussian kernel $\kappa_2:=w_2\kappa_{{\rm G},\sigma}$
for $\sigma:=0.5$ (i.e., $Q=2$).
The parameters and complexities for each algorithm are summarized in Table \ref{tb:A3}.
Fig.~\ref{fig:Nonlinear_dynamics} depicts the results.
The mean dictionary size was:
KNLMS 181.9,
HYPASS 180.0,
MKNLMS 104.8, and
CHYPASS 104.7.
It is consistently observed that
CHYPASS and MKNLMS outperform their single-kernel counterparts with lower complexity.


\subsection{Real Data: Time Series Prediction}
\label{subsec:real_data}

\subsubsection{Experiment B1 - Laser Signal}
\label{subsubsec:laser}
\begin{figure}[t!]
\centering
\subfigure[]{
 \includegraphics[width=7.5cm]{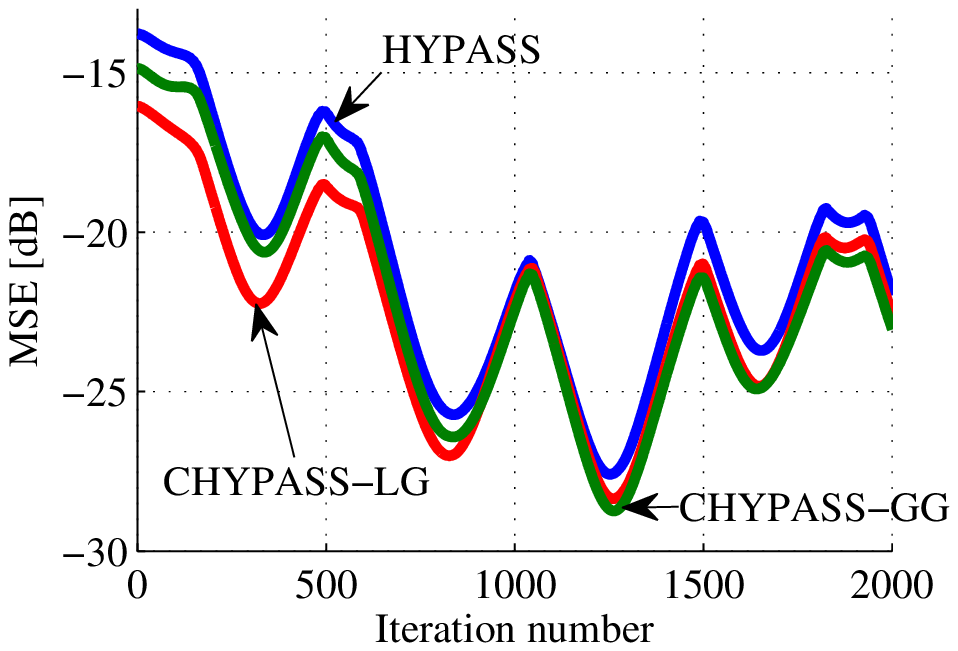}
}
\subfigure[]{
 \includegraphics[width=7.5cm]{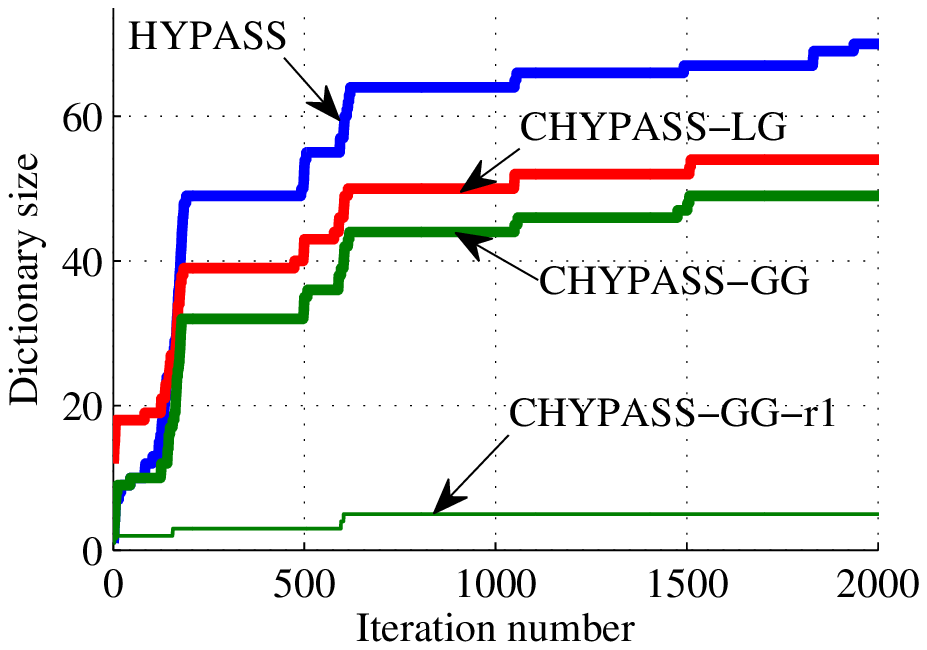}
}
 \caption{Results for Experiment B1: the laser signal.}\vspace{0em}
\label{fig:laser}
\end{figure}

We use the chaotic laser time series from the Santa Fe time series competition
\cite{weigend94} (cf.~\cite{engel04}).
The dataset contains 1,000 samples and we use it twice for learning.
The maximum value of the data is
normalized to one and
is then corrupted by noise $v_n\sim \mathcal{N}(0,1.0\times 10^{-2})$.
We predict each datum $d_n$ with
a collection of past data
$\signal{u}_n:=[d_{n-1}, d_{n-2}\cdots, d_{n-L+1}]^{\sf T}\in\real^L$,
$n\in\Natural$, for $L=10$.
We test two cases of CHYPASS: the linear-Gaussian case
(referred to as CHYPASS-LG)
and the two-Gaussian case (referred to as CHYPASS-GG).
The parameters and complexities for each algorithm are summarized in Table \ref{tb:B1}.
Note that the Gaussian kernel is normalized as in \refeq{eq:Gaussian_kernel}.
In the present case, 
$(\sqrt{2\pi}\sigma_1)^L\approx 0.98\times 10^{4}$ and
$(\sqrt{2\pi}\sigma_2)^L\approx 1.0\times 10^{-3}$.
The unbalance due to the use of large $L$
makes the scale of the autocorrelation matrix of
$\signal{k}_{2,n}$ be much greater than
that of $\signal{k}_{1,n}$ both for CHYPASS and MKNLMS.
This causes extremely slow convergence in terms of those coefficients
associated with $\kappa_1$ and, as a result, 
the performance of the algorithms becomes almost the same as obtained
by the sole use of $\kappa_2$.
To emphasize the effect of $\kappa_1$, a very small value is allocated to
$w_2$ so that $w_1:=1-w_2\approx 1.0$.
Fig.~\ref{fig:laser} depicts the results.
The mean dictionary size was:
HYPASS 62.2,
CHYPASS-LG 49.4, and
CHYPASS-GG 43.9.
It is seen that
CHYPASS outperforms HYPASS with lower complexity.

\begin{table}[t]
\caption{Parameter settings and complexities for Experiment A2.}
\label{tb:A2}
\vspace*{-1em} 
\begin{center}
\begin{tabular}{|c|c|c|c|}
\hline 
&\multicolumn{2}{|c|}{parameter}&complexity\\ \hline
NLMS & \multicolumn{2}{|c|}{$\lambda_n=0.1$}  & {$5$}\\\hline
KNLMS & $\lambda_n=0.1$ &$\delta=0.8$ & {$1232$}\\ \cline{1-1}\cline{3-4}
HYPASS & $\sigma=\sigma_2=0.02$ &$\delta=0.8$, $s_n=1$ & {$829$}\\  \cline{1-1}\cline{3-4}
MKNLMS &$\varepsilon=0.01$  & $\sigma_1=1.0$,  $\delta_1=0.92$ & {$897$}\\ 
(Two-Gaussian)&  & $\delta=0.6$, $w_2=0.1$ & \\ \cline{1-1}\cline{3-4}
CHYPASS&  & $\sigma_1=1.0$, $\delta_1=0.92$  &$609$\\
(Two-Gaussian)&  & $\delta=0.65$, $w_2=0.1$ & \\
&  & $s_n=1$ & \\ \hline
\end{tabular}\vspace*{-1em} 
\end{center}
\end{table}

\subsubsection{Experiment B2 - CO2 Emission Data}
\label{subsubsec:co2}

\begin{figure}[t!]
\centering
\subfigure[]{
 \includegraphics[width=7.5cm]{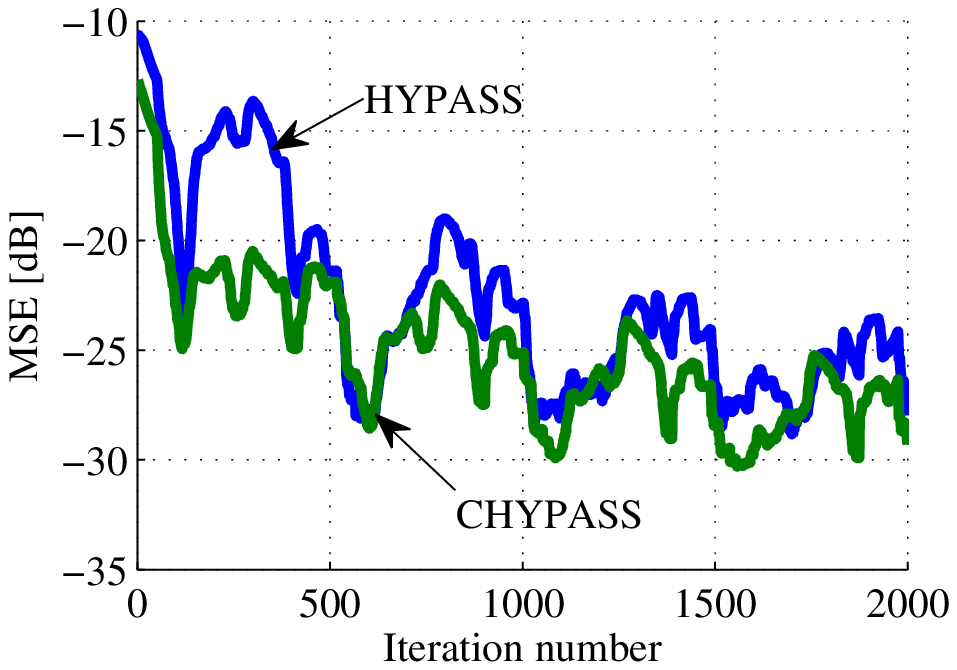}
}
\subfigure[]{
 \includegraphics[width=7.5cm]{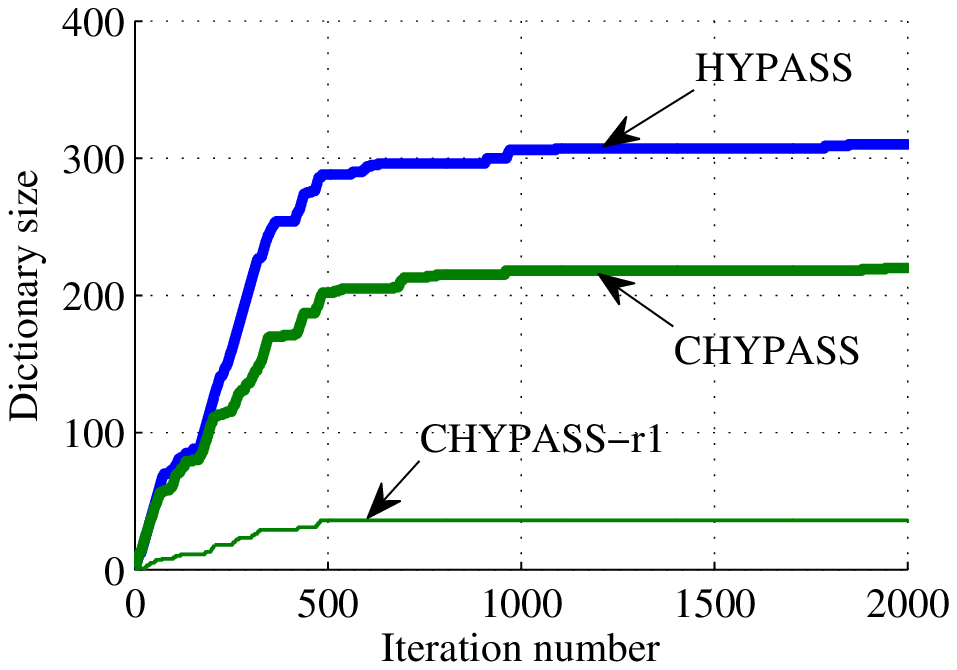}
}
 \caption{Results for Experiment B2: the CO2 emission data.}\vspace{0em}
\label{fig:energy}
\end{figure}

We use a real data of the carbon dioxide emissions from energy consumption in the
industrial sector
during
Jan. 1973 to Jun. 2013,
available at the Data Market website (http://datamarket.com/).
The dataset contains 486 samples and we use it repeatedly for learning.
As in Section \ref{subsubsec:laser},
the maximum value of the data is normalized to one and
is then corrupted by noise $v_n\sim \mathcal{N}(0,1.0\times 10^{-2})$.
Each datum $d_n$ is predicted with
$\signal{u}_n:=[d_{n-1}, d_{n-2}\cdots, d_{n-L+1}]^{\sf T}\in\real^L$,
$n\in\Natural$, for $L=20$.
We test CHYPASS with two-Gaussian kernels and compare its performance
with that of HYPASS.
The parameters and complexities for each algorithm are summarized in
Table \ref{tb:B2}.
Due to the same idea as in Section \ref{subsubsec:laser}, the weight
is designed as
$w_2:=(\sqrt{2\pi}\sigma_2)^L/
(\sqrt{2\pi}\sigma_1)^L\approx 3.5\times 10^{-11}$ and
$w_1:=1-w_2\approx 1.0$.
Fig.~\ref{fig:energy} depicts the results.
The mean dictionary size was:
HYPASS 268.2, and CHYPASS 191.4.
It is seen that
CHYPASS significantly outperforms HYPASS, 
particularly in the initial phase,
with lower complexity.

\begin{table}[t]
\caption{Parameter settings and complexities for Experiment A3.}
\label{tb:A3}
\vspace*{-1em} 
\begin{center}
\begin{tabular}{|c|c|c|c|}
\hline 
&\multicolumn{2}{|c|}{parameter}&complexity\\ \hline
NLMS & \multicolumn{2}{|c|}{$\lambda_n=0.1$}  & 8\\\hline
KNLMS & $\lambda_n=0.1$ & $\delta=0.95$  & 1275 \\ \cline{1-1}\cline{3-4}
HYPASS & $\sigma=0.5$ & $\delta=0.95$, $s_n=1$  & 906 \\ \cline{1-1}\cline{3-4}
MKNLMS & $\varepsilon=0.05$ & $\delta=0.9$, $w_1 = 0.2$  & 729 \\
(Linear-Gaussian) &  &  & \\
 \cline{1-1}\cline{3-4}
CHYPASS&  & $\delta=0.9$, $w_1 = 0.1$ &524\\ 
(Linear-Gaussian)& &$s_n=1$ &\\ \hline
\end{tabular}\vspace*{-1em} 
\end{center}
\end{table}

\subsection{Wrap-up}
\label{subsec:wrapup}

We finally wrap up this experimental section
by reviewing the results from three aspects.

\subsubsection{Multikernel and Single-kernel Approaches}
Comparing the CHYPASS and MKNLMS algorithms with their respective
single-kernel counterparts, we can see that the multikernel approach
exhibit better MSE performances with smaller dictionary sizes.
This indicates that 
the use of multiple kernels would allow 
compact representations of unknown systems as mentioned in 
Section \ref{subsubsec:efficiency}.

\subsubsection{Functional and Parameter-space Approaches}
\label{subsubsec:functional_parameter}
It can be observed that the functional approaches (CHYPASS and
       HYPASS) outperform the parameter-space approaches (MKNLMS
       and KNLMS).
This would be due to the {\it decorrelation effect}
inherent in the functional approach
according to our experimental studies of HYPASS and CHYPASS.
To be specific, we have empirically found that 
the multiplication of the inverse of the kernel matrix (see Lemma \ref{lemma:PMf})
decorrelates the kernelized input vector, 
yielding the improvements of convergence behaviors.

\subsubsection{Efficacy of the Selective Updating Strategy}
In all the experiments, the size $s_n$ of selected subsets is chosen
so that any further increase of $s_n$ does not improve the performance
significantly.
In other words, each functional approach for $s_n\leq 5$ achieves
the best possible performance that is realized by its exact version
(i.e., $s_n=r_n$) which is computationally expensive as explained in
Section \ref{subsec:complexity}.
This clearly shows the efficacy of the selective updating strategy.



\begin{table}[t]
\caption{Parameter settings and complexities for Experiment B1.}
\label{tb:B1}
\vspace*{-1em} 
\begin{center}
\begin{tabular}{|c|c|c|c|}
\hline 
&\multicolumn{2}{|c|}{parameter}&complexity\\ \hline
HYPASS & $\lambda_n=0.1$ &$\delta=0.4$, $s_n=5$ & 971\\  \cline{1-1}\cline{3-4}
CHYPASS& $\sigma=\sigma_2=0.2$ & $\delta=0.2$, $s_n=5$  & 698\\
(Linear-Gaussian)& $\varepsilon=0.05$  & $w_2 = 1.0\times 10^{-3}$ & \\  \cline{1-1}\cline{3-4}
CHYPASS& &$\sigma_1=1$, $\delta_1=0.6$ & 900\\
(Two-Gaussian)&  &  $\delta=0.2$, $s_n=5$ & \\
&  & $w_2 = 1.0\times 10^{-5}$  & \\ \hline
\end{tabular}\vspace*{-1em} 
\end{center}
\end{table}

\begin{table}[t!]
\caption{Parameter settings and complexities for Experiment B2.}
\label{tb:B2}
\vspace*{-1em} 
\begin{center}
\begin{tabular}{|c|c|c|c|}
\hline 
&\multicolumn{2}{|c|}{parameter}&complexity\\ \hline
HYPASS & $\lambda_n=0.1$ & $s_n=5$ & 6331 \\  \cline{1-1}\cline{3-4}
CHYPASS& $\sigma=\sigma_2=0.3$ &$\sigma_1=1$, $\delta_1=0.95$  & 4731 \\
(Two-Gaussian)& $\delta=0.95$  & $s_n=5$ & \\
& $\varepsilon=0.01$  & $w_2 = 3.5\times 10^{-11}$ & \\ \hline
\end{tabular}\vspace*{-1em} 
\end{center}
\end{table}

\clearpage

\section{Concluding Remarks}
\label{sec:conclusion}

We proposed the CHYPASS algorithm 
for the task of estimating/tracking
nonlinear functions which contain multiple components.
The proposed algorithm is based on
iterative orthogonal projections
in the Cartesian product of multiple RKHSs.
The proposed algorithm was derived by reformulating the HYPASS algorithm
in the product space.
In the particular case (including the linear-Gaussian case),
the proposed algorithm can also be regarded as operating
iterative projections in the \sinq{sum} space of the RKHSs.
The numerical examples with three toy models and two real data
demonstrated that the simultaneous use of multiple kernels 
led to a compact representation of the nonlinear functions
and yielded better performance than the single-kernel algorithms.

The two streams of multikernel adaptive filtering and HYPASS
have met and united.
The key idea for the union was presented in the simplest possible way
by focusing on the NLMS-type algorithm.
It is our future work of significant interest
to extend CHYPASS to a more sophisticated one such as
a $\Phi$-PASS type algorithm;
$\Phi$-PASS is based on parallel projection and
thus enjoys better convergence properties than HYPASS
\cite{takizawa_icassp13,takizawa_tsp14}.
Further investigations are definitely required
to verify the practical value of 
the proposed Cartesian-product projection approach
in real-world applications.
It is also our important future issue
to verify the {\it decorrelation effect} mentioned in Section \ref{subsubsec:functional_parameter}
from the theoretical and/or experimental viewpoints.



\appendices


\newcounter{appnum}
\setcounter{appnum}{1}

\setcounter{theorem}{0}
\renewcommand{\thetheorem}{\Alph{appnum}.\arabic{theorem}}

\setcounter{lemma}{0}
\renewcommand{\thelemma}{\Alph{appnum}.\arabic{lemma}}
\setcounter{example}{0}
\renewcommand{\theexample}{\Alph{appnum}.\arabic{example}}
\setcounter{equation}{0}
\renewcommand{\theequation}{\Alph{appnum}.\arabic{equation}}
\setcounter{claim}{0}
\renewcommand{\theclaim}{\Alph{appnum}.\arabic{claim}}
\setcounter{remark}{0}
\renewcommand{\theremark}{\Alph{appnum}.\arabic{remark}}

 \section{Kernel Ridge Regression in Sum Space}
 \label{appendix:krr_sum}

We present a basic theorem for the batch case.
\begin{theorem}[Kernel Ridge Regression in Sum Space]
\label{theorem:krr_sum}
Given a set of finite samples $\{(\signal{u}_j,d_j)\}_{j=1}^r$, 
define a regularized risk functional $R_c(f)$ of $f\in\euclidspace^+$ as
\begin{equation}
 R_{c}(f):=\frac{1}{r}\sum_{j=1}^{r}\left(
f(\signal{u}_j) - d_j
\right)^2 + \eta \norm{f}_{\euclidspace^+}^2, ~~~ \eta > 0.
\label{eq:Rcf}
\end{equation}
Then, the minimizer $f^*:=\argmin_{f\in\euclidspace^+}
R_{c}(f)$ is given by
$f^*=\sum_{j=1}^{r}\alpha_j \kappa(\cdot,\signal{u}_j)$
with
$[\alpha_1,\alpha_2,\cdots,\alpha_r]^{\sf T}
:=(\signal{K}+ \eta r \signal{I})^{-1} [d_1,d_2,\cdots,d_r]^{\sf T}$,
where $\signal{K}\in\real^{r\times r}$ is the kernel matrix whose
$(i,j)$ entry is $\kappa(\signal{u}_i,\signal{u}_j)$.
\end{theorem}

The result in \cite[Theorem 1]{ylxu13}
can be reproduced by applying
Theorem \ref{theorem:krr_sum}
with a weighted norm and its associated kernel
given in Corollary \ref{corollary:weighted_norm}.

\bibliographystyle{IEEEtran}
\bibliography{adptl_rkhs}




%







\end{document}